%% file: TLDR_main_arxiv.tex
\documentclass{article}

    \usepackage[final,nonatbib]{neurips_2025}

\usepackage[utf8]{inputenc} %
\usepackage[T1]{fontenc}    %
\usepackage[colorlinks=true, urlcolor=blue, linkcolor=red, citecolor=green]{hyperref}       %
\usepackage{url}            %
\usepackage{booktabs}       %
\usepackage{amsfonts}       %
\usepackage{nicefrac}       %
\usepackage{microtype}      %
\usepackage{xcolor}         %
\usepackage{amsmath}
\usepackage[%
    style=numeric-comp,
    sorting=none,
    natbib=true,  %
    backend=biber,
    sortcites=true,
    doi=false,
    url=false,
    giveninits=true,
    mincitenames=1,
    maxcitenames=1, %
    minbibnames=3, %
    maxbibnames=4, %
    hyperref
]{biblatex}
\usepackage{graphicx}
\usepackage{cleveref}
\usepackage{multirow}
\usepackage{subcaption}
\bibliography{bibtex/external}

\title{$\mathtt{VIBE}$: Annotation-Free Video-to-Text Information Bottleneck Evaluation for TL;DR}

\author{%
    Shenghui Chen\thanks{Equal contribution (Order determined by coin toss).}, ~~~Po-han Li\footnotemark[1], ~~~Sandeep Chinchali, ~~~Ufuk Topcu\\  
    The University of Texas at Austin\\
    \texttt{\{shenghui.chen, pohanli, sandeepc, utopcu\}@utexas.edu} \\
}

\input{section/_preamble}

\begin{document}
\maketitle

\begin{abstract}
\input{section/0_abstract}
\end{abstract}

\input{figure_latex/system}

\input{section/1_intro}

\input{section/2_related_work}
\input{section/3_preliminaries}
\input{section/4_method}
\input{section/5_user_study}
\input{section/6_conclusion}

\input{section/7_ack}

\newpage
\printbibliography
\newpage
\input{section/appendix}

\end{document}

%% file: section/_preamble.tex
\usepackage{colortbl}
\usepackage[most]{tcolorbox}
\usepackage{wrapfig}
\usepackage{enumitem}

\definecolor{lightgray}{gray}{0.9}
\definecolor{datasetAcolor}{HTML}{A3C4F3} %
\definecolor{datasetBcolor}{HTML}{FBC4AB} %
\definecolor{datasetCcolor}{HTML}{C9E4DE} %

\newcommand{\masked}[1]{\colorbox{gray!20}{#1}}
\usepackage{needspace}

%% file: section/0_abstract.tex
Many decision-making tasks, where both accuracy and efficiency matter, still require human supervision. For example, tasks like traffic officers reviewing hour-long dashcam footage or researchers screening conference videos can benefit from concise summaries that reduce cognitive load and save time. Yet current vision-language models (VLMs) often produce verbose, redundant outputs that hinder task performance. Existing video caption evaluation depends on costly human annotations and overlooks the summaries' utility in downstream tasks.
We address these gaps with \underline{\textbf{V}}ideo-to-text \underline{\textbf{I}}nformation \underline{\textbf{B}}ottleneck \underline{\textbf{E}}valuation (VIBE), an annotation-free method that scores VLM outputs using two metrics: \textit{grounding} (how well the summary aligns with visual content) and \textit{utility} (how informative it is for the task). VIBE selects from randomly sampled VLM outputs by ranking them according to the two scores to support effective human decision-making.
Human studies on \texttt{LearningPaper24}, \texttt{SUTD-TrafficQA}, and \texttt{LongVideoBench} show that summaries selected by VIBE consistently improve performance---boosting task accuracy by up to $61.23\%$ and reducing response time by $75.77\%$ compared to naive VLM summaries or raw video.
\footnote{\href{https://vivianchen98.github.io/VIBE_website/}{Project Website}, \href{https://github.com/UTAustin-SwarmLab/Task-aware-TLDR-Public}{Code}, and \href{https://huggingface.co/datasets/vivianchen98/LearningPaper24/}{LearningPaper24 Dataset}.}

%% file: figure_latex/system.tex
\begin{figure}[h!]
  \centering
  \includegraphics[width=0.95\textwidth, trim=0 0 0 50, clip]{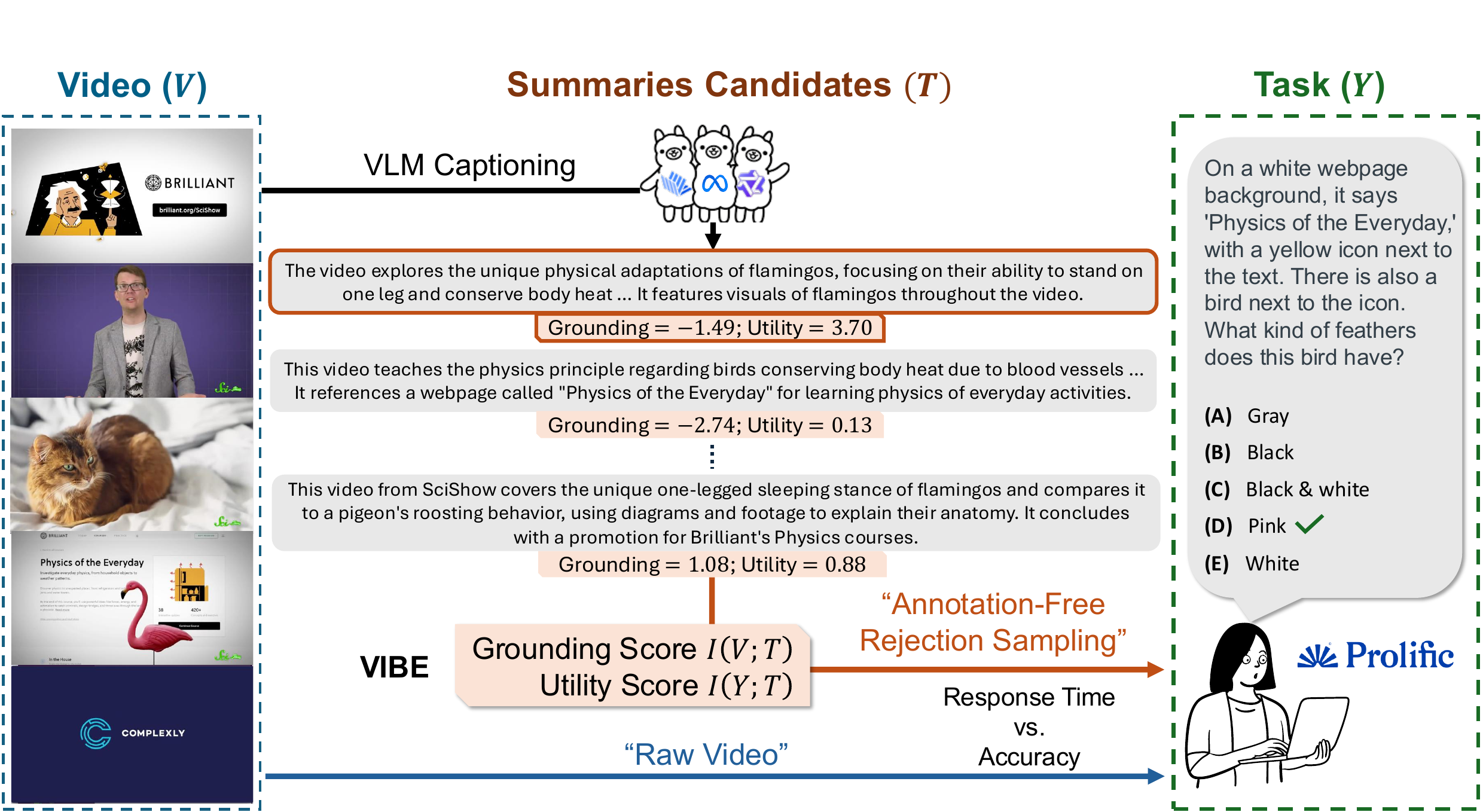}
  \caption{\textbf{VIBE for Video-to-Text Summary Selection.}
    Given a video, a task, and VLM-generated summaries, VIBE ranks the summaries using the proposed grounding and utility scores, which assess video alignment and task relevance. It selects the summary most conducive to helping human users achieve higher task accuracy and lower response time compared to watching the full video.}
  \label{fig:vibe_system}
\end{figure}

%% file: section/1_intro.tex
\section{Introduction}

Efficiently extracting relevant information from extensive video is a major bottleneck for human decision-making, where both accuracy and efficiency matter \cite{HumanAI_Interaction, holzinger2016interactive, humanintheloop_challenge, HOLZINGER202559}.
Tasks demanding human supervision, such as a traffic officer analyzing hours of dashcam footage to determine fault or a researcher distilling key insights from a lengthy oral presentation, are often limited by the time and cognitive load required to process raw video streams. 
In this work, we aim to improve the \textit{quality} and \textit{brevity} of video summaries to boost human task performance compared to existing vision-language model (VLM) outputs and raw video, especially for longer clips where summarization offers greater utility.

Existing video caption evaluation metrics, however, rely heavily on reference-based comparisons to human-annotated summaries \cite{xu2016msr, vatex, auroracap, monfort2021spoken}. These metrics face two main issues.
First, these works require human annotators to watch video clips and write gold-standard captions, which contradicts the goal of reducing human response time and limits generalization to unseen video clips. 
Second, they are oblivious to downstream tasks and fail to measure how well captions support the tasks.

We propose \underline{\textbf{V}}ideo-to-text \underline{\textbf{I}}nformation \underline{\textbf{B}}ottleneck \underline{\textbf{E}}valuation (VIBE), an annotation-free method for selecting task-relevant video summaries without model retraining.
As shown in \Cref{fig:vibe_system}, VIBE defines two metrics---grounding and utility scores---based on the information bottleneck principle \cite{tishby2000informationbottleneckmethod}.
It uses pointwise mutual information to quantify how well a summary reflects video evidence and supports the downstream task.
We leverage next-token prediction in VLMs to access the probability of generating summaries or task answers. By comparing these probabilities with and without key information, we measure how well one modality (text or video) compensates for missing information in the other to assess grounding and task relevance, as shown later in \Cref{fig:vibe}.
Using these scores, VIBE selects the most decision-supportive summary for humans from randomly sampled VLM outputs via annotation-free rejection sampling, which filters candidates without human labels for faster task completion than watching full video clips.

To evaluate VIBE, we conduct between-subjects user studies with $243$ participants across three datasets---\texttt{LearningPaper24} (self-curated), \texttt{LongVideoBench}~\cite{wu2024longvideobench}, and \texttt{SUTD-TrafficQA}~\cite{xu2021sutd}---measuring human performance in terms of accuracy, response time, and inverse efficiency score, the ratio of response time to accuracy \cite{townsend1983stochastic}.
Results show that summaries selected with maximal utility score improve task accuracy by up to $40\%$, while those chosen by maximizing grounding score yield up to $27.6\%$ gains, both on the \texttt{LongVideoBench} dataset.
VIBE-selected summaries also significantly reduce response time compared to raw video.
We also observe a strong positive correlation between utility score and human accuracy, and a strong negative correlation between summary length and response time per word. These patterns highlight the value of concise, relevant information for efficient human decision-making.

\paragraph{Contributions.}
Our contributions are threefold:
(a) We identify the need and propose the problem of annotation-free, task-aware evaluation for video-to-text summarization, improving human response time and accuracy without relying on gold-standard captions.
(b) We propose VIBE, an annotation-free evaluation framework that combines grounding and utility scores of video clips to rank and select high-quality summaries from VLM outputs without requiring retraining.
(c) We demonstrate through user studies that VIBE summaries significantly boost humans' task accuracy by up to $61.23\%$ and reduce response time by up to $75.77\%$ compared to standard VLM outputs. 

\paragraph{Critique and Open Problems.}
This work reframes the video-to-text evaluation problem through the lens of human decision support, offering an annotation-free, scalable alternative to costly reference-based comparisons. 
VIBE’s ability to score and select summaries without training makes it a practical plug-in for both closed- and open-source VLMs.
Looking ahead, VIBE opens several promising directions.
One is exploring the joint optimization of summary generation and selection through self-supervised fine-tuning of VLMs. Another is extending beyond human supervision tasks to investigate whether task-aware captions can improve VLM performance on downstream reasoning tasks.
These directions point to the broader applicability of VIBE beyond evaluating video summarization.

%% file: section/2_related_work.tex
\section{Related Work}
\paragraph{Reference-based Video Caption Evaluation.}
Evaluating video caption quality is crucial for tasks like video question answering \cite{msrvttQA, zhong2022videoquestionansweringdatasets}, text-to-video retrieval \cite{xu2016msr, li2024any2any, omama2025exploiting}, and multimodal language model training. The quality of captions greatly impacts the downstream performance of tasks and models.
While existing benchmarks compare VLM-generated captions to gold-standard references using metrics like ROUGE \cite{lin2004rouge}, BLEU \cite{papineni2002bleu}, and CIDEr \cite{vedantam2015cider}, these are costly to curate and focus solely on captions rather than video context.
In contrast, we propose VIBE, a new evaluation metric that incorporates video context, requires no gold standard, and can be applied to unseen video-caption pairs.
VIBE extends the information-theoretic approach of \cite{jung2024informationtheoretic}, which uses pointwise mutual information to assess news summaries for language model tuning. We adapt it for video captions and examine their effect on human task accuracy and response time.

\paragraph{Human-Centric Evaluation via Response Time and Accuracy.}
In the context of human-agent interaction involving natural language \cite{chen2024human}, traditional human evaluations have focused on assessing summaries based on fluency and informativeness \cite{belz2006comparing, graham2017can}, but these methods are often subjective and hard to scale. To address this, we adopt an extrinsic evaluation approach that measures how summaries affect human performance on downstream tasks \cite{nenkova2011automatic, pu2023summary}. Specifically, we evaluate captions based on human response time and accuracy on multiple-choice questions grounded in video content. To account for the speed-accuracy tradeoff \cite{heitz2014speed}, we also report the inverse efficiency score \cite{townsend1983stochastic}, which normalizes response time by accuracy.

%% file: section/3_preliminaries.tex
\section{Preliminaries}
\paragraph{Mutual Information.}
Our work heavily relies on the concept of mutual information. For the ease of readers, we briefly introduce it here. 
Mathematically, for two random variables, $X,Z$, we can calculate their mutual information:
\begin{equation}
    \mathbf{I}(X;Z) = \mathbf{E} \left[\log \frac{\mathbf{P}(X,Z)}{\mathbf{P}(X)\mathbf{P}(Z)} \right] = \mathbf{E} \left[\log \frac{\mathbf{P}(X|Z)}{\mathbf{P}(X)} \right],
    ~~ \text{(Mutual Information)}
\label{eq:mi}
\end{equation}
where the expectation operator is over the joint probability distribution of $X$ and $Z$.

Intuitively, mutual information measures the reduction in uncertainty about $X$ after observing $Z$. It is zero if and only if $X$ and $Z$ are independent, meaning knowledge of $Z$ provides no information about $X$. Higher mutual information indicates stronger dependency between the two variables. In this work, we leverage mutual information to quantify how much the summaries retain relevant information from the raw video clips and how well they support the target downstream task.

\paragraph{Information Bottleneck.}
The information bottleneck (IB) framework extracts relevant information from input data while compressing irrelevant details.
\citet{tishby2000informationbottleneckmethod} first introduced it to formalize the trade-off between accuracy and complexity in learning systems. 
Later, researchers applied it to domains such as neuroscience \cite{kleinman2023cortical}, natural language processing \cite{west2019bottlesum,zhang2022improving}, and computer vision \cite{1238368,7133169}.
To ensure clarity, we modify the notation from \cite{tishby2000informationbottleneckmethod} to align with our later sections, where $V$ is the raw video input, $T$ is the summary of the video, and $Y$ is the downstream task of our interest.

We now introduce the IB framework, which seeks to learn a representation $T$ that discards sensitive and irrelevant information from the input $V$ while preserving information useful for predicting the target task $Y$.
IB relies on minimizing the mutual information between input $V$ and a compressed representation $T$, while preserving as much information as possible about a target variable $Y$:
\begin{equation}
\min_{\mathbf{P}(t|v)} \underbrace{\mathbf{I}(V;T)}_{\text{compression}} - \beta \underbrace{\mathbf{I}(T;Y)}_{\text{informativeness}}, \quad \text{(Information Bottleneck)}
\label{eq:IB}
\end{equation}
where $\mathbf{I}(\cdot\,;\cdot)$ denotes mutual information, and $\beta\geq 0$ controls the trade-off between compression and prediction. The optimization variable is the conditional distribution $\mathbf{P}(t|v)$, which defines a stochastic encoder that maps $v\in V$ to $t\in T$.
The IB principle in \cref{eq:IB} formalizes the trade-off between compression and prediction. 
Minimizing $\mathbf{I}(V;T)$ encourages stronger compression, while maximizing $\mathbf{I}(T;Y)$ promotes the preservation of task-relevant information.
The parameter $\beta$ balances these two competing objectives: a higher $\beta$ prioritizes retaining more information about $Y$, while a lower $\beta$ encourages stronger compression of $V$.

Researchers use the IB principle to analyze generalization, regularization, and the role of hidden representations of neural networks \cite{kawaguchi2023does, goldfeld2020informationsurvey}, where $V$ is the input of neural networks, $T$ is the output of intermediate layers, and $Y$ is the classification label.
It also drives recent advances in representation learning, where models aim for compact, informative encodings \cite{MMIB, islam2023representation}.

%% file: section/4_method.tex
\section{Video-to-Text Information Bottleneck Evaluation (VIBE)}
\input{figure_latex/vibe}

We now formally describe our video-to-text information bottleneck evaluation (VIBE) method.
Inspired by the IB formulation in \cref{eq:IB}, we adapt its two terms for the purpose of evaluating video-to-text summary quality.
We denote the raw video clips that are fed into the VLMs as $V$, the summaries from the VLM response as $T$, and the task target we care about as $Y$.
For instance, let $V$ be a video of a paper presentation, $T$ be the summary of the presentation, and the task $Y$ be to determine the primary area to see if it is of one's interest.

\paragraph{Grounding Score.} In \cref{eq:IB}, the mutual information $\mathbf{I}(V;T)$ between raw data and representation is the compression term. In our setting, we reinterpret it as the \textit{grounding score}, which quantifies how grounded the textual summary is to the video. The higher this score is, the more closely the summary is anchored to the video.

However, one cannot access the joint and marginal probability distributions over the occurrences of video clips or text summaries.
Therefore, we can only calculate the empirical mutual information $\log \frac{\mathbf{P}(X,Z)}{\mathbf{P}(X)\mathbf{P}(Z)}$ without the expectation operator to approximate the $\mathbf{I}(V;T)$ term. 
Literature refers to the empirical term without the expectation as pointwise mutual information \cite{bouma2009normalized}, which we denote as $\mathbf{I}^P(\cdot;\cdot)$.
Unlike mutual information, which must be non-negative as it is the expectation of pointwise mutual information, pointwise mutual information ranges from $[-\infty, \infty]$.

We now explain the approximation technique in VIBE to calculate the grounding score. Recall from \cref{eq:mi}, we rewrite pointwise mutual information as:
\begin{equation}
    \mathbf{I}^P(V;T) = \log \frac{\mathbf{P}(T|V)}{\mathbf{P}(T)} \\
            \approx \log \frac{\mathbf{P}(T|V, T_{\mathrm{masked}})}{\mathbf{P}(T|T_{\mathrm{masked}})}.
~~ \text{(Grounding Score)}
\label{eq:ground}
\end{equation}
We condition both the numerator and denominator with $T_{\mathrm{masked}}$, a masked version of the text, to approximate $\mathbf{I}^P(V;T)$. The approximation holds if $T_{\mathrm{masked}}$ is independent of both the video $V$ and the original text $T$. Masking key content makes $T_{\mathrm{masked}}$ effectively uninformative and thus independent. We put the detailed derivation in Appendix \ref{app:approx}.
Following \citet{jung2024informationtheoretic}, we use tf-idf \cite{tfidf} to identify keywords and replace them with a "<MASK>" token. This masking helps ensure that \cref{eq:ground} reflects the true dependency between $V$ and $T$, rather than exploiting shortcuts from the text itself. Thus, the grounding score measures how much the TL;DR remains genuinely grounded in the video beyond what the masked text alone would suggest.

To compute \cref{eq:ground}, we run two separate inferences with the same VLM to define the ratio of two conditional probabilities $\mathbf{P}(\cdot|\cdot)$.
The first, $\mathbf{P}(T|V, T_{\mathrm{masked}})$, is the product of next-token prediction probabilities when reconstructing the masked text given both the video and the masked input.
The second, $\mathbf{P}(T|T_{\mathrm{masked}})$, is computed similarly but without providing the video.
Only by conditioning on the masked input can we leverage the VLM's next-token prediction probabilities to estimate these likelihoods.
By dividing these two conditional probabilities and taking the logarithm, we approximate the pointwise mutual information, allowing us to estimate the grounding score.

\paragraph{Utility Score.} The second term in \cref{eq:IB}, the informativeness term, is the mutual information $\mathbf{I}(T;Y)$ between the representation and the target task.
We refer to it as the \textit{utility score}, which measures how useful the summary is for solving the downstream task. A higher utility score indicates that the summary preserves more information relevant to predicting or completing the target task.
Similar to the grounding score, directly computing $\mathbf{I}(T;Y)$ is intractable because we cannot access the true probability distributions. Again, we approximate it using the pointwise mutual information with conditional probabilities:
\begin{equation} \mathbf{I}^P(T;Y) = \log \frac{\mathbf{P}(Y|T)}{\mathbf{P}(Y)} \approx  \log \frac{\mathbf{P}(Y|T, V_{\mathrm{masked}})}{\mathbf{P}(Y|V_{\mathrm{masked}})}.
~~ \text{(Utility Score)}
\label{eq:utility} 
\end{equation}
Similar to the grounding score in \cref{eq:ground}, $V_{\mathrm{masked}}$ denotes video clips with key information removed to reduce their direct informativeness about both the summary $T$ and the task label $Y$. To compute the utility score in \cref{eq:utility}, we compare two model inferences: one conditioned on both the summary $T$ and the masked video $V_{\mathrm{masked}}$, and the other on $V_{\mathrm{masked}}$ alone. Since both use the same masked visual input, any improvement in predicting $Y$ must come from the information provided by $T$.
Intuitively, if a summary is useful, it should compensate for the missing content in the masked video and help recover task-relevant information. Thus, the utility score quantifies how well the summary restores information necessary for predicting the task, directly measuring the summary’s contribution to downstream decision-making.

\input{figure_latex/vibe_scatter}

\paragraph{Rejection Sampling over Grounding and Utility Scores.}
We now describe how VIBE optimizes grounding and utility scores to select more informative and task-relevant video-to-text summaries for humans, as shown in \Cref{fig:vibe_system}.
Given multiple summaries $T$ sampled from the VLM's output distribution $P(T|V)$, VIBE selects the candidate that maximizes a weighted sum of the two pointwise mutual information terms:
\begin{equation}
{\arg\max}_{T \sim P(T|V)} \alpha \; \mathbf{I}^P(V;T) + \beta \; \mathbf{I}^P(T;Y).
~~ \text{(TL;DR Selection with VIBE)}
\label{eq:vibe}
\end{equation}
Here, $\alpha,\beta$ are hyperparameters controlling the trade-off between grounding and utility in \cref{eq:vibe}. A higher $\alpha$ emphasizes alignment with video, favoring summaries that faithfully reflect the video, while a higher $\beta$ prioritizes task relevance, selecting summaries that improve downstream performance.

Our formulation differs from traditional IB, where compression serves as a regularizer during model training to mitigate overfitting. In contrast, VIBE does not train a model---it selects a summary. Overfitting is therefore not a concern, and maximizing both mutual information terms is desirable to produce summaries that are both useful and faithful.
To this end, we consider $\alpha, \beta \geq 0$ to jointly maximize both scores. 
Then, to explore potential trade-offs between the two scores, we perform a convex combination sweep over $\alpha \in \{0,\, 0.05,\, 0.1,\, \ldots,\, 1.0\}$, with $\beta = 1 - \alpha$. Each $(\alpha, \beta)$ pair is applied consistently across all datasets. 
This linear scalarization technique, commonly used in multi-objective optimization \cite{boyd2004convex}, produces the convex (``Pareto curve") shown in \Cref{fig:ib_pareto}, empirically revealing the inherent trade-off between grounding and utility scores.

Our results based on grounding and utility scores across sampled summaries from multiple datasets reveal a consistent trade-off between the two objectives. As shown in \Cref{fig:ib_pareto}, summaries selected by VIBE with varying $\alpha$ and $\beta$ form a Pareto front, capturing the inherent trade-off between the two VIBE scores. In contrast, randomly sampled summaries (\textit{Naive VLM}) consistently fall inside this Pareto curve across datasets, indicating suboptimal grounding and utility.
We observe this pattern when evaluating the two VIBE scores of VLM-generated summaries on tasks from \texttt{LearningPaper24}, \texttt{SUTD-TrafficQA}, and \texttt{LongVideoBench}.
Summaries with high grounding scores tend to mirror the video content closely but often include redundant or irrelevant details for the task. In contrast, summaries with high utility scores boost task accuracy and response efficiency but may overlook important visual context. This trade-off reflects the core tension in the IB framework. By tuning $\alpha$ and $\beta$, VIBE adjusts this balance to match the needs of different downstream applications. We detail the experimental settings in the next section.

As shown empirically in the experiments, higher utility scores correlate with higher human task accuracy.
Thus, one can use VIBE to select summaries with higher utility scores and present only the most informative summaries to humans, filtering out less useful ones.
Notably, calculating the utility score requires access to task labels $Y$, but not gold-standard, human-annotated labels for summary $T$ as in previous works.
When task labels are unavailable, VIBE can select summaries based on the highest grounding score, an unsupervised measure of alignment between the video and the generated summaries.
Although the grounding score alone yields lower human performance than the utility score, our user study shows it still improves task accuracy compared to unfiltered VLM summaries.

\paragraph{From Evaluation to Real-World Use.}
VIBE evaluates the quality of summaries only from the VLM perspective, but its connection to human decision-making performance remains unanswered solely by the formulation.
To bridge this gap, we conduct user studies measuring how different summary qualities, as evaluated by VIBE, impact human decision-making performance in \Cref{sec:user_study}.
VIBE relies solely on text generation and next-token probability access, using black-box access to VLMs like the OpenAI API \cite{openai_python}. It enables VIBE to work with both closed- and open-source VLMs and scale well to real-world scenarios.

%% file: figure_latex/vibe.tex
\begin{figure}[ht]
  \centering
  \includegraphics[width=\textwidth, trim=0 0 0 25, clip]{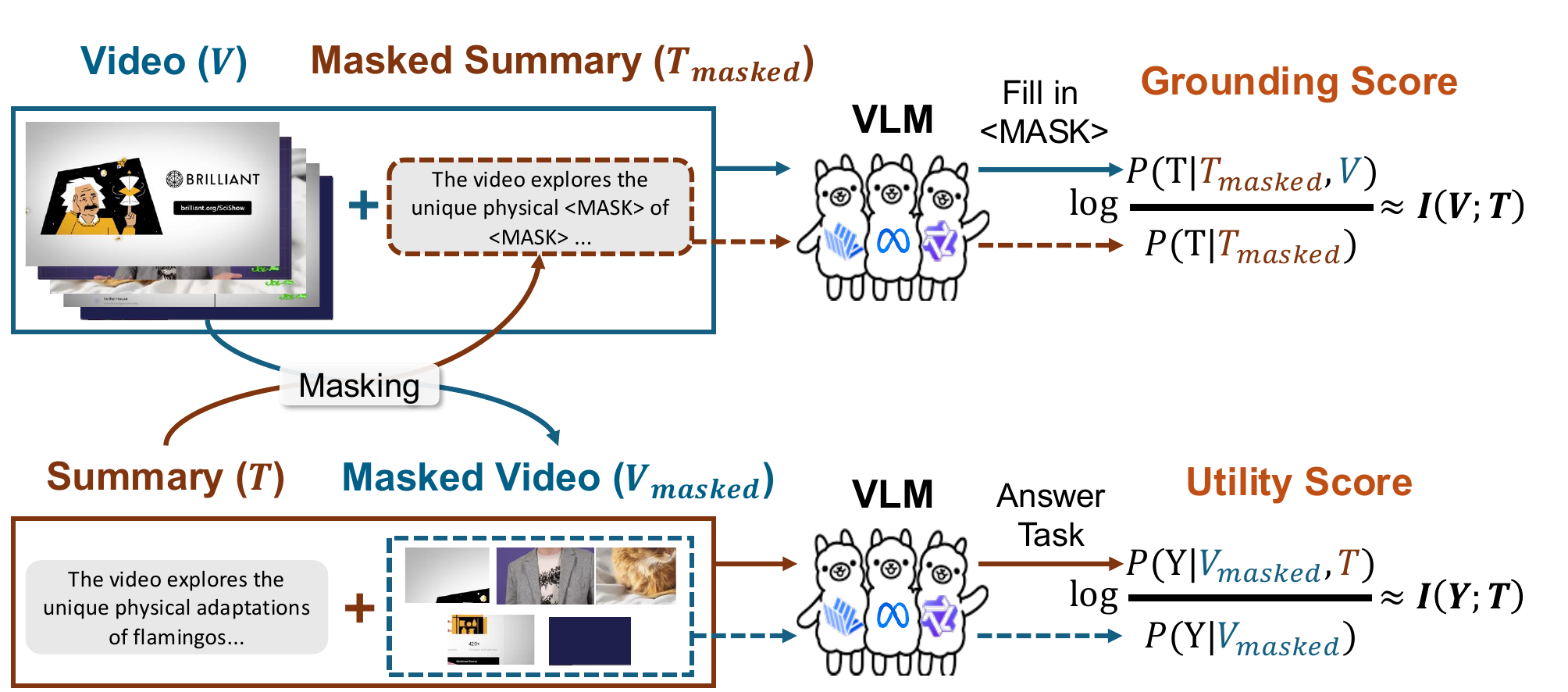}
  \caption{\textbf{Computing VIBE Scores via Masked Inference.}
    VIBE estimates grounding and utility scores using the next-token prediction mechanism of VLMs. The grounding score measures how well the video helps reconstruct a masked summary, while the utility score captures how much the summary improves task prediction given a masked video.
    }
  \label{fig:vibe}
\end{figure}

%% file: figure_latex/vibe_scatter.tex
\begin{figure}[b]
    \vspace{-1em}
    \centering

    \includegraphics[width=0.8\linewidth]{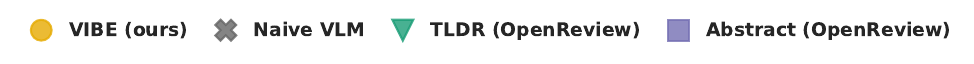} \\
    \vspace{0.5em}

    \begin{minipage}[b]{0.32\linewidth}
        \centering
        \includegraphics[width=\linewidth]{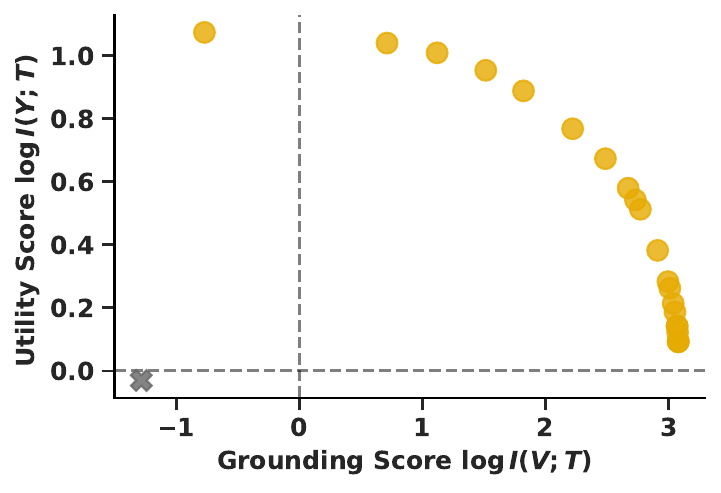}
        \subcaption{\texttt{LongVideoBench}}
    \end{minipage}
    \hfill
    \begin{minipage}[b]{0.32\linewidth}
        \centering
        \includegraphics[width=\linewidth]{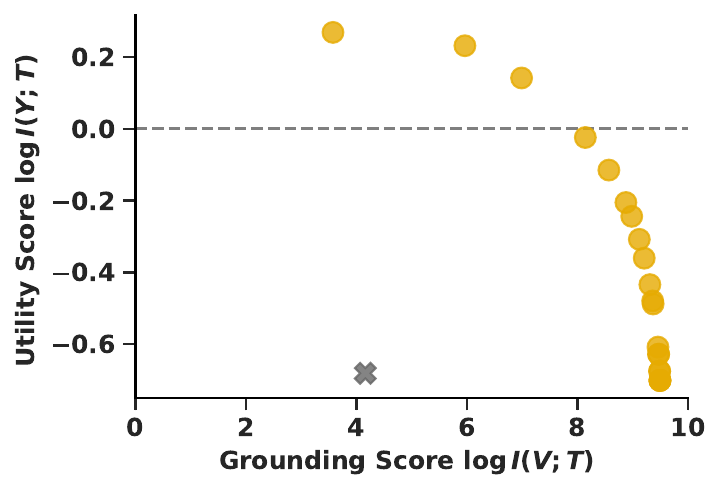}
        \subcaption{\texttt{SUTD-TrafficQA}}
    \end{minipage}
    \hfill
    \begin{minipage}[b]{0.32\linewidth}
        \centering
        \includegraphics[width=\linewidth]{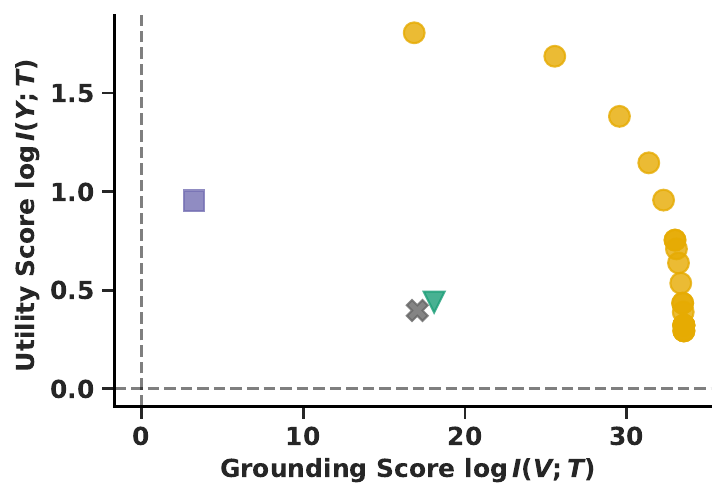}
        \subcaption{\texttt{LearningPaper24}}
    \end{minipage}
    
    \caption{\textbf{Pareto front of VIBE.} Summaries selected by VIBE form a Pareto frontier across different $(\alpha, \beta)$ per \Cref{eq:vibe}, demonstrating optimal trade-offs between grounding and utility. In contrast, Naive VLM summaries, author-written TLDRs, and abstracts from OpenReview in \texttt{LearningPaper24} fall inside the frontier, indicating suboptimal scores.}
    \label{fig:ib_pareto}
\end{figure}

%% file: section/5_user_study.tex
\section{Experiment}\label{sec:user_study}
We conduct user studies across three diverse datasets to evaluate how VIBE-generated summaries support human decision-making. The study measures participants' accuracy and response time in answering questions about video content.
Here, we use Qwen2.5-VL-72B-AWQ \cite{bai2025qwen25vltechnicalreport} to evaluate VIBE for all datasets.
A representative qualitative example from the study is provided in \Cref{appendix:qualitative}.
We also show that VIBE works across various VLMs through ablation studies.

\subsection{Datasets}
For evaluation, we select three datasets varying in domain, duration, and task nature. 

\paragraph{LearningPaper24}
We introduce \texttt{LearningPaper24}, a curated dataset of 2,287 video presentations from \textsc{ICLR 2024} and \textsc{NeurIPS 2024}, filtered based on key criteria: a valid OpenReview ID, an accessible SlidesLive video, an author-provided TL;DR and abstract, and a clearly defined primary area. Details on the curation process and dataset statistics are provided in \Cref{appendix:learningpaper24}.
The task associated with this dataset is to identify the primary area of each paper from $12$ options. See instruction details in \Cref{appendix:study_interfaces}.

\paragraph{LongVideoBench \& SUTD-TrafficQA}
\texttt{LongVideoBench} \cite{wu2024longvideobench} features long instructional clips with QA pairs for extended reasoning, while \texttt{SUTD-TrafficQA} \cite{xu2021sutd} contains short traffic clips with multiple-choice questions on causal and temporal understanding. Participants answer one question per each \texttt{LongVideoBench} clip; four questions per each \texttt{SUTD-TrafficQA} clip.

\paragraph{Preprocessing}
In all datasets, we mask text by removing keywords with high tf-idf scores. For \texttt{LearningPaper24}, which features slide videos, we extract and mask keywords in the slide using EasyOCR \cite{easyocr}. For the other two datasets, we apply random $1/16$ cropping to all video frames.
We use a subset of each dataset for VIBE evaluation and user studies (details provided in \Cref{appendix:selection}).

\subsection{User Study Design}\label{subsec:study_design}
For each dataset, 10 video stimuli with multiple-choice questions are shown in randomized order. We adopt a \textit{between-subjects design} where participants are assigned to one of the four conditions listed below, and the format in which the stimulus is presented varies by the condition. 

\paragraph{Independent Variables.}\label{para:IV}
Participants are assigned to one of four conditions. In the \textbf{Video Only} condition, they watch the original video without text. In the remaining conditions, they view only a VLM-generated summary: a randomly selected VLM summary (\textbf{Naive}), the top-ranked summary from $k$ response candidates\footnote{$k=5$ in this study, and the $k$ responses are generated with various temperatures.} by utility score (\textbf{Max-U}) or by grounding score (\textbf{Max-G}).

\paragraph{Dependent Measures.}
We report three metrics: \textbf{accuracy}, the proportion of correct responses across 10 stimuli; \textbf{response time}, the time (in seconds) spent reading or watching each stimulus and answering its corresponding questions; and \textbf{inverse efficiency score (IES)}, the ratio of response time to accuracy, to account for speed–accuracy trade-offs \cite{townsend1983stochastic}.

\paragraph{Hypotheses.}
We evaluate the following hypotheses:
(\textbf{H1}) Participants in the Max-U and Max-G will achieve higher accuracy than those in Naive VLM. (\textbf{H2}) Participants in the Max-U and Max-G will respond more quickly than those in the Video Only. (\textbf{H3}) Max-U and Max-G will yield lower (i.e., more efficient) IES scores than Video Only, reflecting a better speed–accuracy trade-off.

\paragraph{Participants.}
We recruit $243$ participants across three datasets: $92$ for \texttt{LearningPaper24}, $82$ for \texttt{SUTD-TrafficQA}, and $69$ for \texttt{LongVideoBench}. For \texttt{LearningPaper24}, participants are primarily CS graduate students or prescreened degree holders on Prolific~\cite{palan2018prolific}; participants for the other datasets are recruited generally. The average age is $37.59 \pm 11.06$ years, with a gender distribution of $63.37\%$ male, $35.80\%$ female, and $0.82\%$ non-binary. Further details are provided in \Cref{appendix:participants}.

\subsection{User Study Results and Analysis}
We assess statistical significance using the independent t-test, reporting the t-statistic $t(df)$ (with degrees of freedom $df$), the significance level $p$, and the effect size measured by Cohen’s $d$.

\begin{table}
    \centering
    \small
    \setlength{\tabcolsep}{3pt}
    \resizebox{\textwidth}{!}{
    \begin{tabular}{ccc|cccc}
        \toprule \rowcolor{white} \multirow{2}{*}{\textbf{Dataset}}          & \multirow{2}{*}{\textbf{Duration (s)}}              & \multirow{2}{*}{\textbf{Metric}}  &  \multicolumn{4}{c}{\textbf{IV Conditions}} \\
         &                           &                  & \textit{Video Only}                          & \textit{Naive}                      & \textit{Max-G (ours)} & \textit{Max-U (ours)}                         \\
        \midrule

\multirow{3}{*}{\texttt{LearningPaper24}} & \multirow{3}{*}{250--325} & Acc $\uparrow$   & 23.50 $\pm$ 10.62                                          & 28.42 $\pm$ 12.68                   & 35.00 $\pm$ 7.91      & \textbf{37.89 $\pm$ 10.04} \\
                                                            &                           & RT $\downarrow$  & 192.73 $\pm$ 108.01                                        & 47.48 $\pm$ 20.31                   & 61.13 $\pm$ 39.33     & \textbf{46.69 $\pm$ 25.06} \\
                                                            &                           & IES $\downarrow$ & 9.85 $\pm$ 6.63                                            & 1.81 $\pm$ 1.51                     & 1.85 $\pm$ 1.20       & \textbf{1.28 $\pm$ 0.66}   \\
        \midrule

\multirow{3}{*}{\texttt{LongVideoBench}}  & \multirow{3}{*}{221--489} & Acc $\uparrow$   & \textbf{74.44 $\pm$ 14.23}              & 46.43 $\pm$ 13.94                   & 59.23 $\pm$ 9.17      & 65.00 $\pm$ 15.47                             \\
                                                            &                           & RT $\downarrow$  & 202.35 $\pm$ 87.26                                         & 86.50 $\pm$ 63.33                   & 71.93 $\pm$ 35.96     & \textbf{65.86 $\pm$ 41.34} \\
                                                            &                           & IES $\downarrow$ & 2.93 $\pm$ 1.48                                            & 1.83 $\pm$ 1.01                     & 1.25 $\pm$ 0.67       & \textbf{1.14 $\pm$ 0.80}   \\
        \midrule

\multirow{3}{*}{\texttt{SUTD-TrafficQA}}  & \multirow{3}{*}{2--10}    & Acc $\uparrow$   & 82.86 $\pm$ 5.08                                           & 76.96 $\pm$ 6.89                    & 80.47 $\pm$ 3.21      & \textbf{84.81 $\pm$ 4.65}  \\
                                                            &                           & RT $\downarrow$  & \textbf{48.63 $\pm$ 21.30}              & 79.46 $\pm$ 25.77                   & 80.97 $\pm$ 41.11     & 137.27 $\pm$ 81.64                            \\
                                                            &                           & IES $\downarrow$ & \textbf{0.59 $\pm$ 0.27}                & 1.05 $\pm$ 0.37                     & 1.01 $\pm$ 0.52       & 1.66 $\pm$ 1.02                               \\
        \bottomrule
    \end{tabular}
    }
    \vspace{0.5em}
    \caption{Human performance (mean $\pm$ standard deviation) across three datasets under different IV conditions (detailed in \Cref{para:IV}). Metrics: Accuracy (Acc, \%, higher is better), Response Time (RT, seconds, lower is better), and Inverse Efficiency Score (IES = RT/Acc, lower is better). Bolded values indicate the best performance among the IV conditions for each metric.}
    \label{tab:comparison}
    \vspace{-1.5em}
\end{table}

\input{figure_latex/main_result_plot}

\paragraph{On H1 (Accuracy).}
As shown in \Cref{tab:comparison} and (a1, b1) of \Cref{fig:qualitative_main} (with Max-U in yellow and Max-G in purple above Naive VLM in pink), both Max-U and Max-G consistently outperform the Naive VLM across all datasets, confirming \textbf{H1}. Max-U achieves the largest gains, with statistically significant improvements over Naive VLM in \texttt{LearningPaper24} ($t(36)=2.486, p=0.009, d=0.806$), \texttt{LongVideoBench} ($t(26)=3.215, p=0.002, d=1.261$), and \texttt{SUTD-TrafficQA} ($t(25)=3.311, p=0.001, d=1.325$), likely due to its focus on task-relevant content that helps users quickly locate key information. However, computing the utility score requires task labels, which may limit its general applicability and scalability. 
Max-G, while yielding smaller gains, still significantly outperforms the baseline (e.g., \texttt{LearningPaper24}: $t(33)=1.750, p=0.045, d=0.594$; \texttt{LongVideoBench}: $t(25)=2.691, p=0.006, d=1.077$; \texttt{SUTD-TrafficQA}: $t(28)=1.759, p=0.045, d=0.667$). Crucially, calculating the grounding score is fully self-supervised and does not require task labels, making it an alternative that still delivers meaningful accuracy gains.

\paragraph{On H2 (Response Time).}
\Cref{tab:comparison} shows that both Max-U and Max-G significantly reduce response times compared to the Video baseline in \texttt{LearningPaper24} ($t(37) = -5.599, p < .001, d = -1.794$) and \texttt{LongVideoBench} ($t(34) = -4.503, p < .001, d = -1.510$). 
However, this trend does not extend to \texttt{SUTD-TrafficQA}, where response time shows no consistent improvement. It is not surprising, as the significantly shorter video durations (2–10 seconds, compared to over 3 minutes in the other datasets) inherently limit the benefits of summarization. 
These results align with the expectation that VIBE is most effective in reducing cognitive load for long video clips.

\paragraph{On H3 (IES).}
As shown in \Cref{tab:comparison}, Max-U significantly outperform the Video Only baseline in \texttt{LearningPaper24} ($t(37) = -5.458, p < .001, d = -1.749$) and \texttt{LongVideoBench} ($t(21) = -3.592, p = .001, d = -1.617$), with Max-G showing similar gains (\texttt{LearningPaper24}: $t(34) = -4.631$, $p < .001$, $d = -1.553$; \texttt{LongVideoBench}: $t(20) = -3.421$, $p = .001$, $d = -1.567$). 
Furthermore, Max-U also outperform Naive VLM ($t(26) = -1.940, p = .032, d = -0.761$) in \texttt{LongVideoBench}.
These results highlight VIBE’s effectiveness in settings where processing full video clips is costly. 
In contrast, \texttt{SUTD-TrafficQA} shows limited IES improvement: despite Max-U's accuracy improvements over Naive VLM and Video Only, the brevity of the clips and the visual nature of fine-grained actions reduce the benefit of text-based summaries.

\paragraph{Correlation Analyses.}
We explore the correlation between utility score, grounding score, word count, and human performance (accuracy and response time) using Spearman’s rank coefficient \cite{spearman1961proof}, which is robust to outliers and non-linear relationships. Full plots are provided in \Cref{appendix:more_user_study}. 
Two consistent trends hold across datasets:
First, utility score positively correlates with accuracy: strongly in \texttt{LongVideoBench} ($r_s = 0.684$, $p < .001$), and moderately in \texttt{LearningPaper24} ($r_s = 0.399$, $p = .011$) and \texttt{SUTD-TrafficQA} ($r_s = 0.463$, $p = .004$). 
Second, word count is strongly negatively correlated with response time per word (\texttt{LongVideoBench}: $r_s = -0.761$, \texttt{LearningPaper24}: $r_s = -0.684$, \texttt{SUTD-TrafficQA}: $r_s = -0.658$, all $p < .001$), implying that longer summaries lead to faster processing per word, possibly due to reduced engagement and more shallow reading.

\paragraph{Key Takeaways.}
Max-U and Max-G outperform Naive VLMs in accuracy, response time, and inverse efficiency score, with Max-U providing the most significant improvements, especially for longer video clips. Max-G, though offering smaller gains, remains a self-supervised alternative that improves accuracy without task labels.
The benefits of summarization are less pronounced in shorter video clips, like those in \texttt{SUTD-TrafficQA}, where brief durations limit gains.
Correlation analyses show the VIBE utility scores align with higher accuracy, while longer summaries tend to increase unit response time, highlighting the importance of concise, relevant information.

\subsection{Ablation over VLM Variants}
\label{sec:var_vlm}
VIBE scores can inherit bias from the underlying VLMs due to their training data and conditional probability estimates used in mutual information computation. To assess robustness under such model-induced bias, we compare VLMs of varying size and source. As shown in \Cref{app:ablation}, different backbones produce consistent IB score trends and scales. This consistency suggests that lightweight models can efficiently verify outputs from larger ones. To further reduce bias, one can evaluate a single summary using multiple VLMs---an ensemble-style strategy aligned with mixture-of-experts \cite{ong2025routellm, zhou2022mixture} and model selection techniques \cite{li_online_2024}. 
This reflects the intuition that good outputs are often hard to generate, but easy to verify.

\paragraph{Limitations.}
Despite being training-free, VIBE requires multiple VLM inferences for grounding score evaluation, with the number of masked tokens influencing inference steps. Its pointwise mutual information estimate may be biased by two main factors:
(a) the text and video masking strategy, and (b) the conditional probability modeling of the VLMs. 
For (a), the tf-idf score we use to select keywords for text masking requires hyperparameter tuning, and the optimal masking strategy remains an open question.
For (b), VLMs introduce bias through their training data and modeling assumptions in computing \cref{eq:vibe}. Using separate models for generation and evaluation, as discussed in \Cref{sec:var_vlm}, can help mitigate this bias.

%% file: figure_latex/main_result_plot.tex
\begin{figure}
    \centering
    \begin{subfigure}{\linewidth}
        \centering
        \includegraphics[width=0.9\linewidth, trim=75 10 80 80, clip]{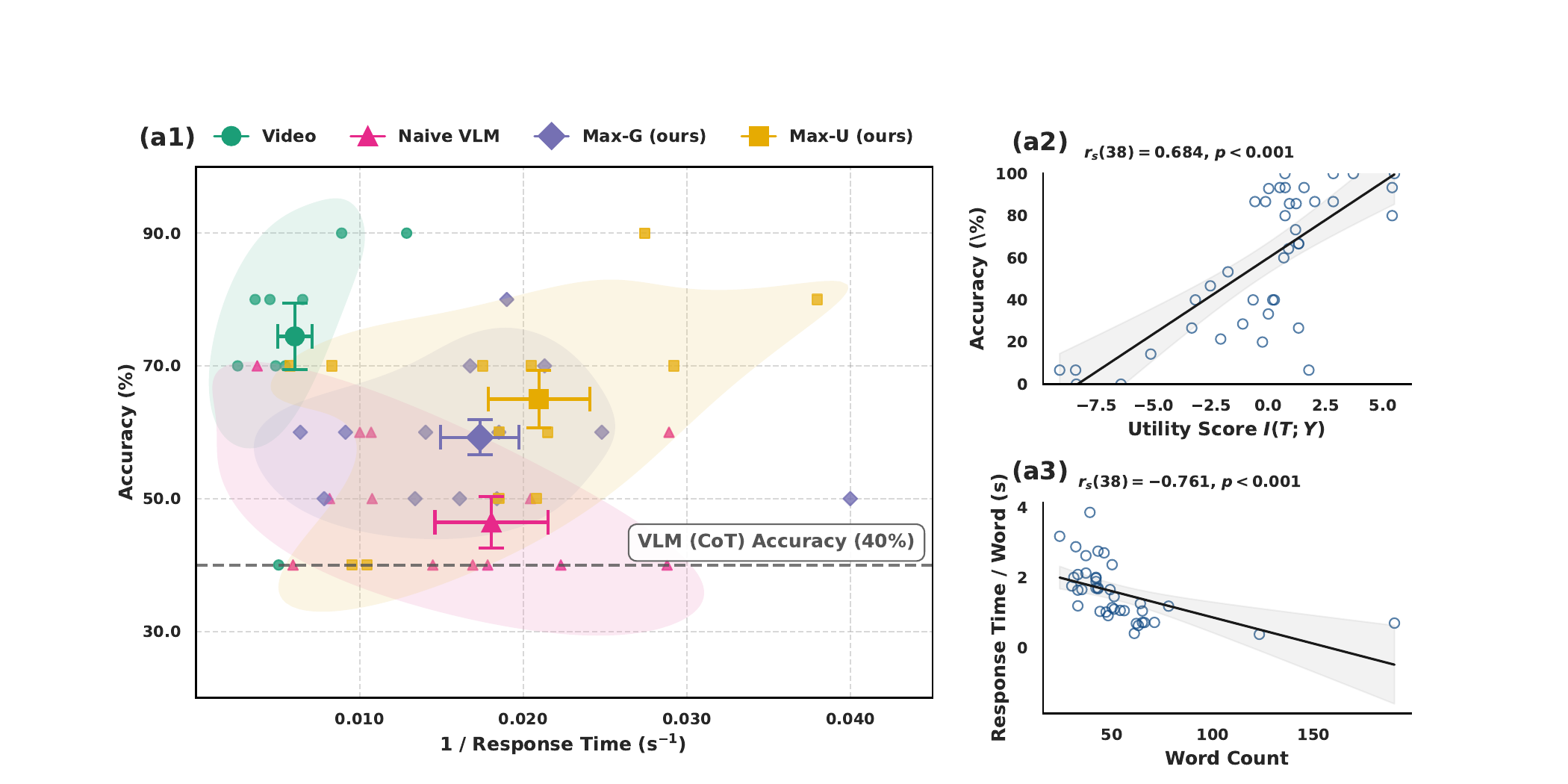}
        \caption{\texttt{LongVideoBench}}
        \label{fig:longvideobench_main}
    \end{subfigure}
    \begin{subfigure}{\linewidth}
        \centering
        \includegraphics[width=0.9\linewidth, trim=75 10 80 60, clip]{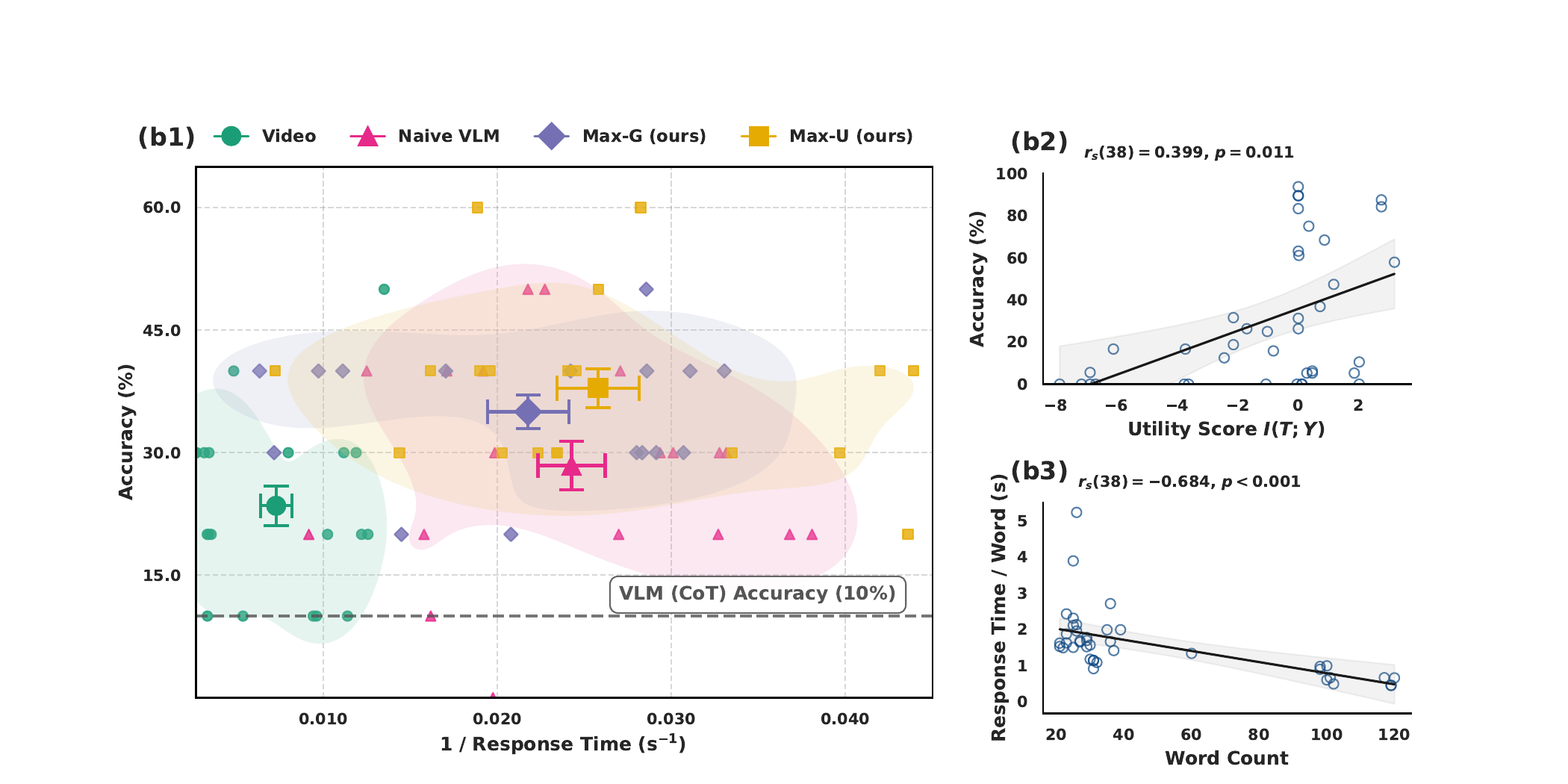}
        \caption{\texttt{LearningPaper24}}
        \label{fig:paper_main}
    \end{subfigure}
    
    \caption{\textbf{(a1, b1)} Accuracy versus inverse response time. Each point represents an individual participant; large markers indicate group means with standard error of the mean. Shaded areas denote 2D kernel density estimates (threshold = $0.45$). 
    \textbf{(a2, b2)} Correlation between accuracy and utility score.
    \textbf{(a3, b3)} Correlation between response time per word and word count.}
    \label{fig:qualitative_main}
    \vspace{-1em}
\end{figure}

%% file: section/6_conclusion.tex
\section{Conclusion}
In this work, we introduce VIBE, an annotation-free framework for evaluating and selecting video-to-text summaries to support human decision-making. Unlike traditional caption evaluation metrics that rely on human-written references, VIBE uses information-theoretic scores—utility and grounding—to assess how well a summary supports a downstream task and aligns with video evidence. This enables scalable, task-aware summary selection without the need for retraining or annotations.
Through a large-scale user study spanning three diverse datasets, we show that summaries selected by VIBE significantly improve both task accuracy and response time, especially for longer video clips where information overload is more likely.

%% file: section/7_ack.tex
\section*{Acknowledgement}
This work was supported in part by the National Science Foundation grants No. CNS-1836900, 2148186, the Office of Naval Research (ONR) under Grant No. N00014-22-1-2254, N00014-24-1-2097, the Defense Advanced Research Projects Agency (DARPA) contract FA8750-23-C-1018, and DARPA ANSR: RTXCW2231110.
Any opinions, findings, and conclusions or recommendations expressed in this material are those of the authors and do not necessarily reflect the views of the National Science Foundation.

%% file: section/appendix.tex
\appendix
\section*{\LARGE Appendix}

\section{Qualitative Result}\label{appendix:qualitative}

To illustrate differences across summary generation methods, we present a qualitative example. 
In \Cref{tab:condition_responses}, each condition---Naive, Max-U, Max-G, and CoT---produces a distinct response based on the same video input (keyframe thumbnails shown in \Cref{fig:qualitative_example_thumbnails}).
For better comparison, we also provide the original TL;DR and abstract from the authors on OpenReview in \Cref{tab:openreview}. We highlight key terms derived from OpenReview metadata and tf-idf analysis by masking them in gray.

\input{figure_latex/qualitative_example_thumbnails}

\begin{table}[ht]
    \centering
    \footnotesize
    \begin{tabular}{lp{0.85\textwidth}}
    \toprule
    \textbf{Type} & \textbf{Paper Summary} \\
    \midrule
    \textbf{TL;DR} & We study the \masked{challenges} that arise when learning \masked{reward} functions with human \masked{feedback} from \masked{partial} observations\\
    \midrule
    \textbf{Abstract} & Past analyses of reinforcement learning from human \masked{feedback} (\masked{RLHF}) assume that the human evaluators fully observe the environment. What happens when human \masked{feedback} is based only on \masked{partial} observations? We formally define two failure cases: deceptive inflation and overjustification. Modeling the human as Boltzmann-rational w.r.t. a belief over trajectories, we \masked{prove} conditions under which \masked{RLHF} is guaranteed to \masked{result} in \masked{policies} that deceptively inflate their performance, overjustify their behavior to make an impression, or both. Under the new assumption that the human's \masked{partial} \masked{observability} is \masked{known} and accounted for, we then analyze how much information the \masked{feedback} process provides about the return function. We show that sometimes, the human's \masked{feedback} determines the return function uniquely up to an additive constant, but in other realistic cases, there is irreducible ambiguity. We propose exploratory \masked{research} directions to help tackle these \masked{challenges} and experimentally validate both the theoretical concerns and potential mitigations, and caution against blindly applying \masked{RLHF} in partially observable \masked{settings}.\\
    \bottomrule
    \end{tabular}
    \vspace{0.5em}
    \caption{Author-provided TL;DR and abstract on OpenReview.}
    \label{tab:openreview}
\end{table}

\begin{table}[ht!]
    \centering
    \footnotesize
    \begin{tabular}{lp{0.85\textwidth}}
    \toprule
    \textbf{Condition} & \textbf{Summary Response} \\
    \midrule
    \textbf{Naive} & This paper explores the \masked{challenges} of \masked{partial} \masked{observability} in \masked{reinforcement} learning from \masked{human} feedback, \masked{focusing} on deceptive inflation and overjustification in \masked{AI} \masked{systems}. \\
    \midrule
    \textbf{Max-U} & This research \masked{investigates} \masked{challenges} in ensuring \masked{AI} safety and \masked{addressing} \masked{partial} \masked{observability} through advanced frameworks in \masked{human}-\masked{machine} interactions and \masked{reinforcement} learning. \\
    \midrule
    \textbf{Max-G} & The \masked{presentation} discusses the \masked{challenges} of \masked{partial} \masked{observability} in \masked{reinforcement} learning from \masked{human} feedback, \masked{focusing} on deceptive inflation and overjustification in \masked{AI} \masked{systems}. \\
    \midrule
    \textbf{CoT} & The slides discuss challenges in reinforcement learning from human \masked{feedback}, particularly focusing on \masked{partial} \masked{observability} and the \masked{issues} of deceptive inflation and overjustification. The content delves into the framework of Markov Decision Processes (MDPs) and the human choice model, which are \masked{central} to reinforcement learning. The presentation also addresses \masked{safety} problems and \masked{alignment} concerns, which are \masked{critical} in the context of human-\masked{AI} interaction. Given the focus on reinforcement learning and the interaction between \masked{AI} and human evaluators, the most suitable area for this paper is one that encompasses both reinforcement learning and the \masked{ethical} \masked{considerations} of human-\masked{AI} interaction. \\
    \bottomrule
    \end{tabular}
    \vspace{0.5em}
    \caption{Responses under each condition for the qualitative example, keywords from OpenReview and tf-idf masked in grey.}
    \label{tab:condition_responses}
\end{table}

\newpage
\section{Ablation Study on Evaluating with Various VLMs}
\label{app:ablation}
\input{figure_latex/ablation_study}
We conduct an ablation study to examine how different VLM variants affect the
reliability of VIBE. Specifically, we compare QwenVL-2.5-72B-AWQ
\cite{bai2025qwen25vltechnicalreport} (used in the main paper) with InternVL2.5-8B-MPO
\cite{chen2025expandingperformanceboundariesopensource} and InternVL3-38B \cite{zhu2025internvl3exploringadvancedtraining}.
These models vary in size, architecture, and pretraining sources. We apply VIBE to
the same set of summaries to compute their utility and grounding scores. Results
in \Cref{fig:ablation} show consistent VIBE score trends across models, indicating
that smaller or alternate VLMs yield similar evaluation signals while reducing inference
cost and potential bias. Notably, various VLMs give similar ranges of both utility
and grounding scores. The consistency across models further supports the
generalizability of VIBE as a training-free framework.

\section{\texttt{LearningPaper24} Dataset Curation}
\label{appendix:learningpaper24}

We source the dataset from the public
\href{https://github.com/papercopilot/paperlists}{PaperCopilot PaperLists} repository, which indexes all accepted papers at \textsc{ICLR 2024} and \textsc{NeurIPS 2024}. 
To ensure data quality and completeness, we apply the following filtering criteria: (1) the paper must have a valid OpenReview ID to guarantee access to full metadata and public reviews; (2) the associated SlidesLive presentation video must be accessible and playable, verified via a headless browser; (3) a TL;DR summary must be present on OpenReview; and (4) a primary area must be specified. After applying these filters, we retain 2,287 entries from an initial set of 2,556.

\begin{figure}[ht]
    \centering
    \includegraphics[width=0.85\linewidth]{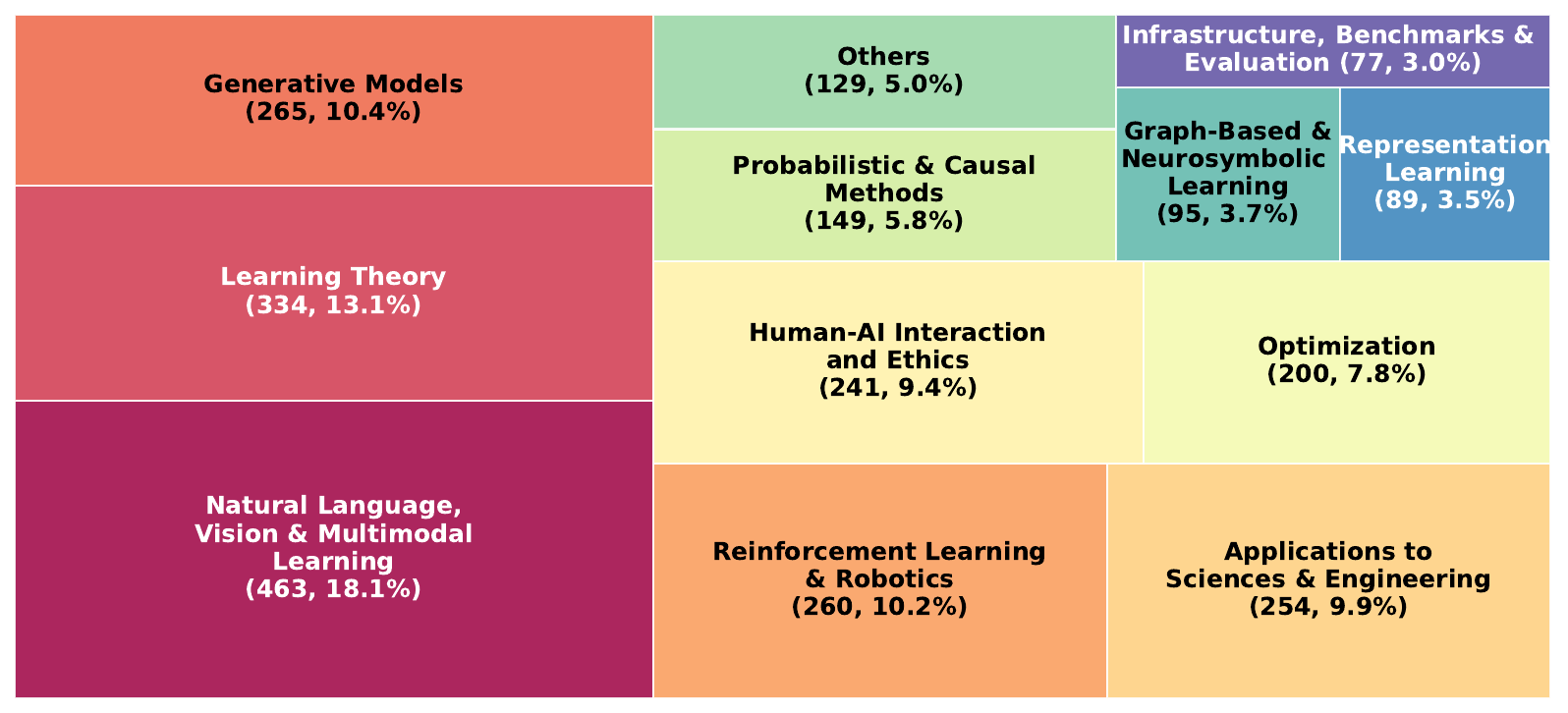}
    \caption{Distribution of papers in \texttt{LearningPaper24} by consolidated primary area.}
    \label{fig:paper_dataset_categories}
\end{figure}

OpenReview's primary areas are often too specific or overlapping, so we remap them into 12 broader and semantically coherent categories. For example, ``self-supervised learning” and ``representation learning for vision” become \textit{Representation Learning}, while ``probabilistic methods” and ``causal inference” are grouped as \textit{Probabilistic \& Causal Methods}. The final taxonomy is shown in \Cref{fig:paper_dataset_categories}.
The final dataset includes the following required fields: OpenReview ID, SlidesLive talk ID and URL, TL;DR, abstract, and primary area.

\section{VIBE Masking and Evaluation}
\label{appendix:selection}
We now report how we select data to calculate the VIBE score in \Cref{fig:ib_pareto}. We further select 10 stimuli for our user study.

\paragraph{LearningPaper24}
We randomly sample 80 papers each from ICLR 2024 and NeurIPS 2024 following the curation process in \Cref{appendix:learningpaper24}. For each presentation, we generate 5 VLM responses and 1 CoT response. We compute the tf-idf score separately for each of the six response corpora. We discard words and phrases (n-grams with range 1–3) appearing in over 10\% of responses and retain only those with tf-idf scores above 0.0025. We also leverage the keywords of papers provided by the authors on OpenReview for masking. We uniformly sample 20 frames from each video as VLM input, with or without random cropping applied.

\paragraph{SUTD-TrafficQA}
We randomly sample 100 video clips from the dataset. For each video clip, we generate 5 VLM responses and 1 CoT response. All responses, including CoT, form the corpus for tf-idf computation. We discard words and phrases (n-grams with range 1–3) appearing in over 30\% of responses and retain only those with tf-idf scores above 0.01. We uniformly select 20 frames from each video as the input of VLM with or without random cropping. We uniformly sample 20 frames from each video as VLM input, with or without random cropping applied.

\paragraph{LongVideoBench}
We sample 150 video clips, each 30 to 500 seconds long, from categories \texttt{["E2O", "E2E", "O3O", "S2A", "S2E", "S2O", "SOS"]}. For each clip, we generate 5 VLM responses and 1 CoT response. All responses form the tf-idf corpus. We discard words and phrases (n-grams with range 1–3) appearing in more than 50\% of responses and keep only those with tf-idf scores above 0.006. We uniformly select 20 frames from each video as the input of VLM with or without random cropping. We use the LongVideoBench package to obtain 32 frames, which is also uniformly sampling the video, from each video as VLM input, with or without random cropping applied.

\section{User Study Instruction and Interfaces}
\label{appendix:study_interfaces}

All participants provide informed consent before participation and are compensated at a rate consistent with Prolific guidelines and institutional standards. The user interface is presented in \Cref{fig:interfaces_side_by_side}.
Below are the instructions provided to participants for each dataset. Text in parentheses indicates variations between video and text conditions.
\input{figure_latex/user_study_interface}

\begin{tcolorbox}
    [ colframe=datasetBcolor!70!black, colback=datasetBcolor!10!white, colbacktitle=datasetBcolor!70!black,
    coltitle=white, title=\textbf{Instructions for \texttt{LongVideoBench}}, sharp
    corners=south, boxrule=0.8mm, width=\textwidth, enlarge left by=0mm, enlarge
    right by=0mm ] 
    \footnotesize 
    In this study, you will (watch/read) \textbf{10} \textbf{(short
    videos/short summaries)} featuring a variety of everyday content. Topics may
    include:
    \begin{itemize}[itemsep=2pt, parsep=0pt]
        \item Cooking
        \item STEM education
        \item Art and creativity
        \item Daily vlogs and lifestyle scenes
    \end{itemize}

    After each (video/summary), you will answer \textbf{one multiple-choice
    question} to assess your recall and understanding of the content.

    Please \textbf{avoid leaving the page idle}, as we are also measuring your
    response time.
\end{tcolorbox}

\begin{tcolorbox}
    [ colframe=datasetCcolor!70!black, colback=datasetCcolor!10!white, colbacktitle=datasetCcolor!70!black,
    coltitle=white, title=\textbf{Instructions for \texttt{SUTD-TrafficQA}}, sharp
    corners=south, boxrule=0.8mm, width=\textwidth, enlarge left by=0mm, enlarge
    right by=0mm ] 
    \footnotesize
    In this study, you'll (watch/read) \textbf{10 (short videos/sets
    of descriptions)} of real traffic videos, which may involve an accident. \\

    After each (video/description), you'll answer \textbf{4 multiple-choice questions}. These
    questions may involve:
    \begin{itemize}[itemsep=2pt, parsep=0pt]
        \item Understanding what happened in the video
        \item Considering what might have happened under different conditions
        \item Reflecting on your own interpretation or reasoning
    \end{itemize}

    Please \textbf{avoid leaving the page idle}, as we are also measuring your response
    time.
\end{tcolorbox}

\begin{tcolorbox}
    [ colframe=datasetAcolor!70!black, colback=datasetAcolor!10!white, colbacktitle=datasetAcolor!70!black,
    coltitle=white, title=\textbf{Instructions for \texttt{LearningPaper24}}, sharp
    corners=south, boxrule=0.8mm, width=\textwidth, enlarge left by=0mm, enlarge
    right by=0mm ] 
    \footnotesize
    In this study, you will (watch/read) \textbf{10} \textbf{(short
    research talk videos/paper summaries)}. 

    For each (video/summary), select the \textbf{research area} you think it belongs to from:

    \begin{itemize}[itemsep=2pt, parsep=0pt]
        \item[(A)] Learning Theory
        \item[(B)] Representation Learning
        \item[(C)] Generative Models
        \item[(D)] Optimization
        \item[(E)] Probabilistic \& Causal Methods
        \item[(F)] Reinforcement Learning \& Robotics
        \item[(G)] Graph-Based \& Neurosymbolic Learning
        \item[(H)] Natural Language, Vision \& Multimodal Learning
        \item[(I)] Human-AI Interaction and Ethics (Privacy, Fairness \& Safety)
        \item[(J)] Applications to Sciences \& Engineering
        \item[(K)] Infrastructure, Benchmarks \& Evaluation
        \item[(L)] Others
    \end{itemize}
    Please \textbf{avoid leaving the page idle}, as we are also measuring your
    response time.
\end{tcolorbox}

\section{Participant Recruitment and Demographics}
\label{appendix:participants}
\Cref{tab:participant_demographics} summarizes demographic information for participants recruited across the three datasets. 

\begin{table}[ht]
    \centering
    \small
    \begin{tabular}{lcll}
        \toprule \textbf{Dataset} & \textbf{\# Participants} & \textbf{Gender Distribution}                    & \textbf{Age (years)} \\
        \midrule \texttt{LearningPaper24}  & 92                       & 70.65\% male, 28.26\% female, 1.09\% non-binary & $32.86 \pm 8.03$     \\
        \texttt{SUTD-TrafficQA}            & 82                       & 59.76\% male, 40.24\% female                    & $40.67 \pm 12.69$    \\
        \texttt{LongVideoBench}            & 69                       & 57.97\% male, 40.58\% female, 1.45\% non-binary & $40.22 \pm 12.40$    \\
        \midrule Total            & 243                      & 63.37\% male, 35.80\% female, 0.82\% non-binary & $37.59 \pm 11.06$    \\
        \bottomrule
    \end{tabular}
    \vspace{0.5em}
    \caption{Demographic breakdown of participants by dataset.}
    \label{tab:participant_demographics}
    \vspace{-1em}
\end{table}

Participants were assigned to IV conditions as shown in \Cref{tab:condition_distribution}. 
In addition to the four primary conditions---Max-U, Max-G, Naive, and Video Only (see \Cref{subsec:study_design})---each dataset also includes a group evaluating Chain-of-Thought (CoT) responses.
CoT responses are excluded from the scatter plots in the main text due to their fundamentally different generation process: unlike other conditions, CoT has access to answer options of the task, making direct comparisons unfair and potentially misleading.
However, we include CoT responses in the correlation analyses, as their grounding and utility scores remain valid for assessing alignment with human performance.

\begin{table}[ht]
    \centering
    \small
    \setlength{\tabcolsep}{6pt}
    \begin{tabular}{lccccc}
        \toprule \textbf{Dataset} & \textbf{Max-U} & \textbf{Max-G} & \textbf{Naive} & \textbf{Video Only} & \textbf{CoT} \\
        \midrule \texttt{LearningPaper24}  & 19             & 16             & 19             & 20             & 18           \\
        \texttt{SUTD-TrafficQA}            & 14             & 17             & 15             & 16             & 20           \\
        \texttt{LongVideoBench}            & 15             & 14             & 15             & 10             & 15           \\
        \bottomrule
    \end{tabular}
    \vspace{0.5em}
    \caption{Number of participants per IV condition across datasets.}
    \label{tab:condition_distribution}
    \vspace{-1em}
\end{table}

\section{Additional User Study Plots}
\label{appendix:more_user_study}
\begin{figure}[t]
    \centering

    \begin{subfigure}[t]{0.9\linewidth}
        \centering
        \includegraphics[width=\linewidth]{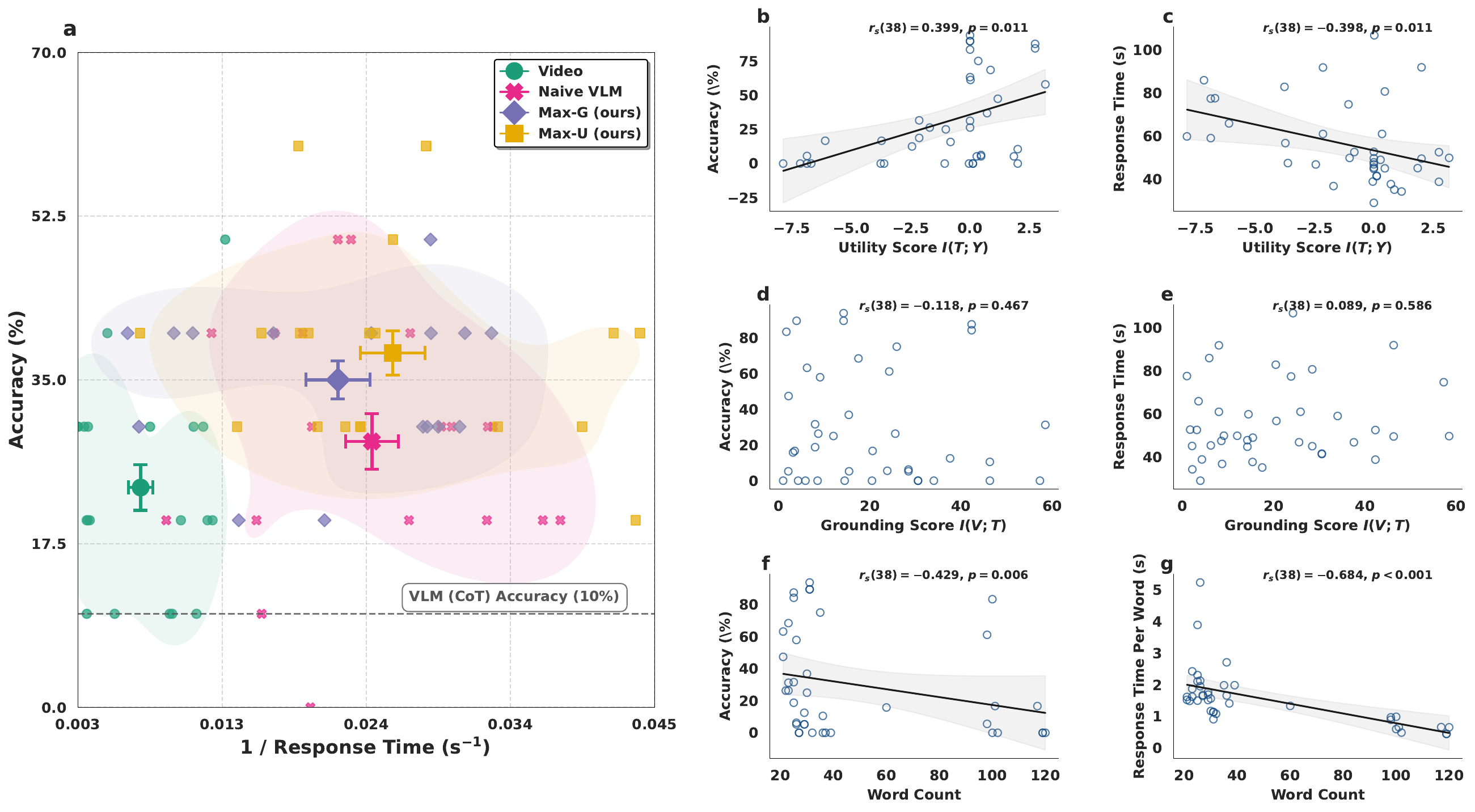}
        \caption{\texttt{LearningPaper24}}
        \label{fig:paper_combined_plot}
    \end{subfigure}

    \vspace{1em}

    \begin{subfigure}[t]{0.9\linewidth}
        \centering
        \includegraphics[width=\linewidth]{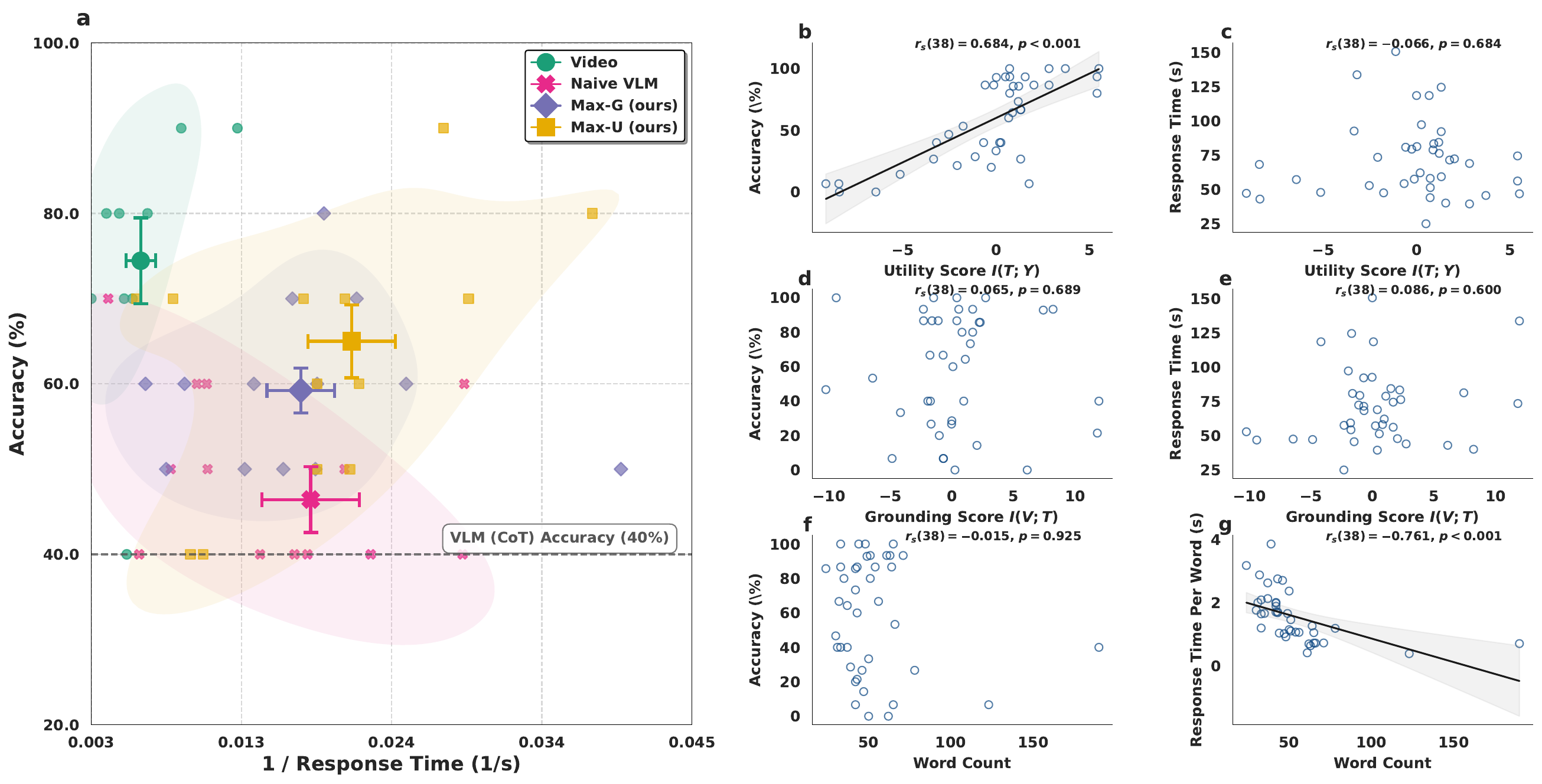}
        \caption{\texttt{LongVideoBench}}
        \label{fig:longvideobench_combined_plot}
    \end{subfigure}

    \vspace{1em}

    \begin{subfigure}[t]{0.9\linewidth}
        \centering
        \includegraphics[width=\linewidth]{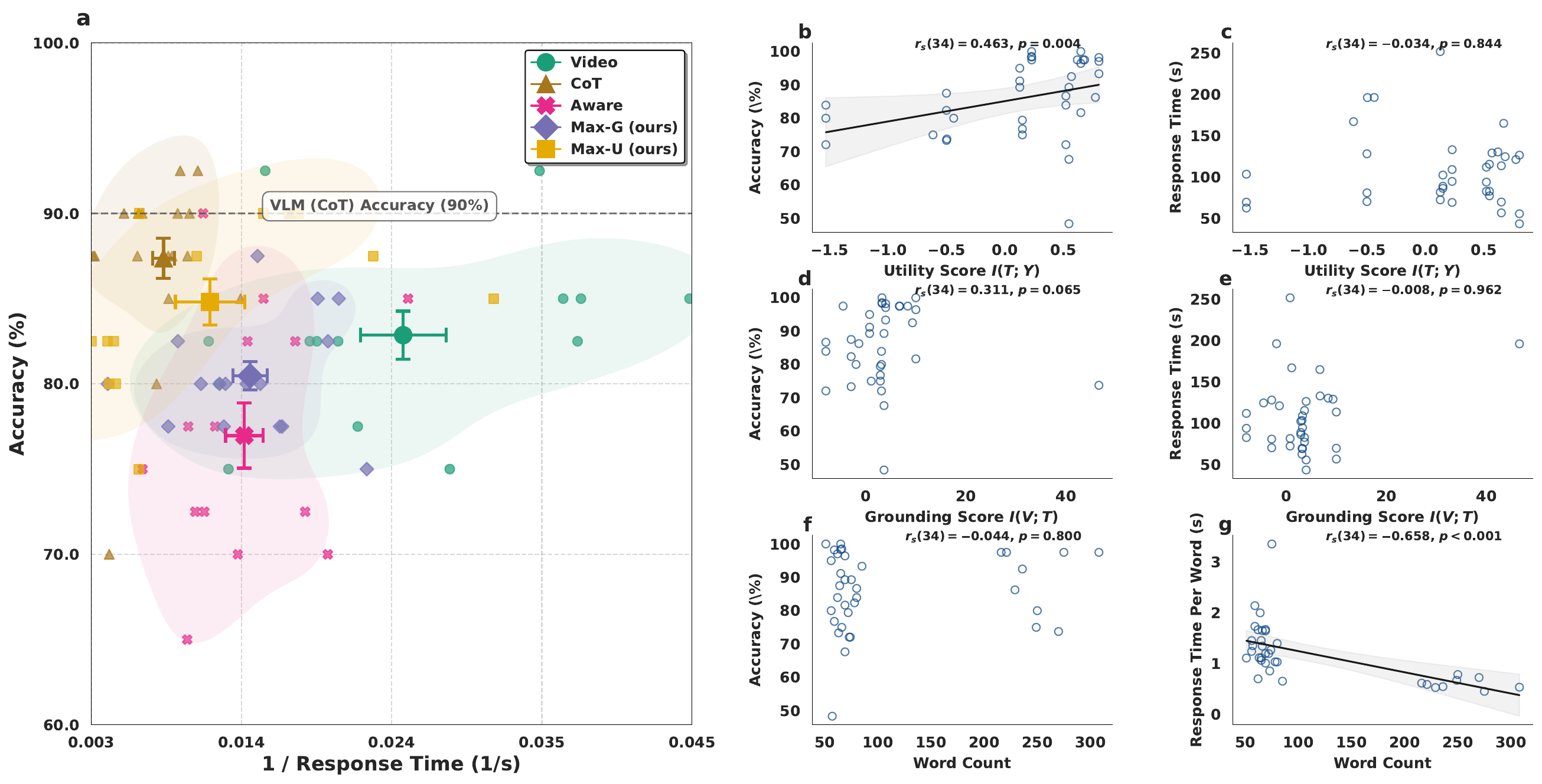}
        \caption{\texttt{SUTD-TrafficQA}}
        \label{fig:traffic_combined_plot}
    \end{subfigure}

    \caption{Scatter plots and Spearman correlations across all datasets. Trendlines shown for $p < 0.05$.}
    \label{fig:combined_correlation_appendix}
\end{figure}

\Cref{fig:combined_correlation_appendix} presents the full set of scatter plots and Spearman correlation results for all datasets. The left panels show scatter plots of accuracy versus inverse response time, while the right panels display correlations between accuracy and utility score, response time and utility score, accuracy and grounding score, response time and grounding score, accuracy and word count, and response time per word and word count.

\section{Computation Resource}
\label{app:comp}
All experiments were conducted on four NVIDIA RTX 6000 Ada GPUs (48GB VRAM) using the vLLM backend. The system was equipped with an Intel(R) Xeon(R) Gold 6346 CPU @ 3.10GHz, featuring 64 cores (x86\_64, 64-bit). This setup handled all models—Qwen-2.5-72B-AWQ, InternVL-2.5-8B-MPO, and InternVL-3-38B—efficiently for both generation and evaluation.

\section{Derivation of Mutual Information Approximation}
\label{app:approx}
We derive the case where the approximation holds under the stated assumption, starting from the right-hand side of Eq. \ref{eq:ground} and Eq. \ref{eq:utility}. Since both approximations follow nearly identical steps, we present the derivation of grounding score as an example:
$$
\log \frac{\mathbf{P}(T \mid V, T_{\mathrm{masked}})}{\mathbf{P}(T \mid T_{\mathrm{masked}})}.
$$

Since $T_{\mathrm{masked}}$ is a masked sentence with all keywords removed, we assume it contains no information and is independent from any other random variables. Thus, we can rewrite the expression as:

$$
\begin{aligned}
\log \frac{\mathbf{P}(T \mid V, T_{\mathrm{masked}})}{\mathbf{P}(T \mid T_{\mathrm{masked}})}
&= \log \frac{\mathbf{P}(T, V, T_{\mathrm{masked}}) / \mathbf{P}(V, T_{\mathrm{masked}})}{\mathbf{P}(T, T_{\mathrm{masked}}) / \mathbf{P}(T_{\mathrm{masked}})} \\
&= \log \frac{\mathbf{P}(T, V) \cdot \mathbf{P}(T_{\mathrm{masked}}) / (\mathbf{P}(V) \cdot \mathbf{P}(T_{\mathrm{masked}}))}{\mathbf{P}(T) \cdot \mathbf{P}(T_{\mathrm{masked}}) / \mathbf{P}(T_{\mathrm{masked}})} \\
&= \log \frac{\mathbf{P}(T, V)/\mathbf{P}(V)}{\mathbf{P}(T)} \\
&= \log \frac{\mathbf{P}(T \mid V)}{\mathbf{P}(T)}.
\end{aligned}
$$

All steps follow from Bayes’ theorem and the assumption of independence.

%% file: figure_latex/qualitative_example_thumbnails.tex
\begin{figure}[ht]
    \centering

    \includegraphics[width=0.24\linewidth]{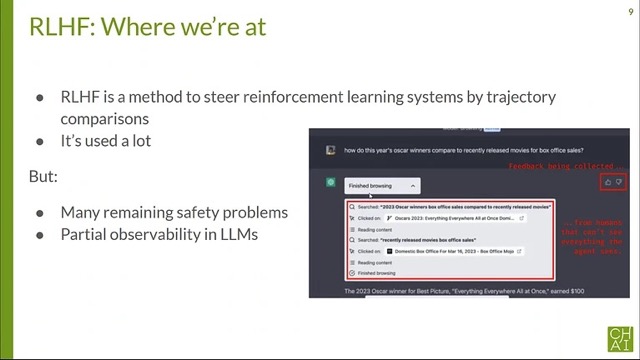}
    \includegraphics[width=0.24\linewidth]{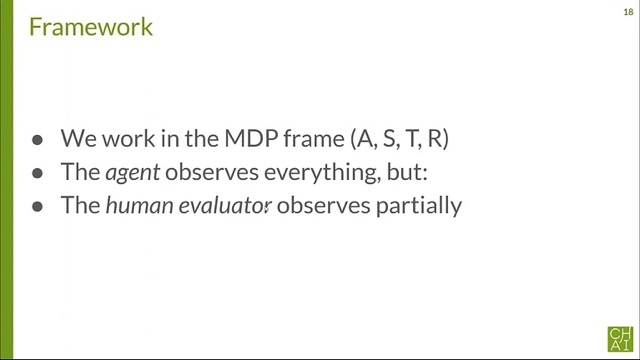}
    \includegraphics[width=0.24\linewidth]{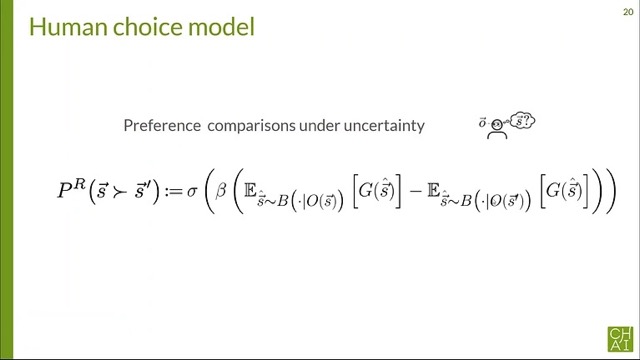}
    \includegraphics[width=0.24\linewidth]{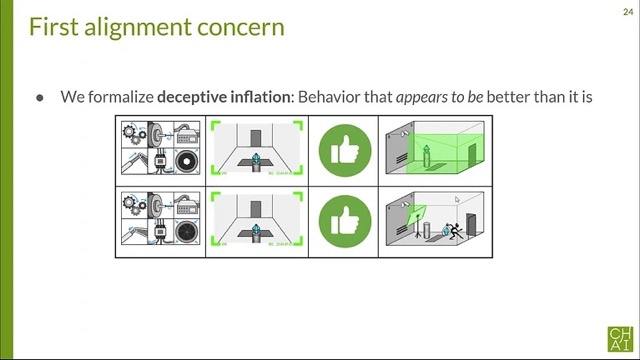} \\[4pt]
    \includegraphics[width=0.24\linewidth]{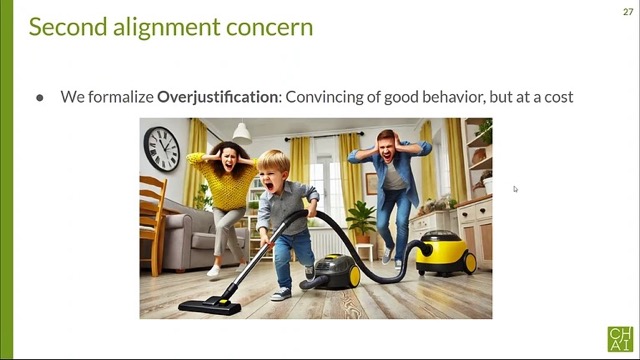}
    \includegraphics[width=0.24\linewidth]{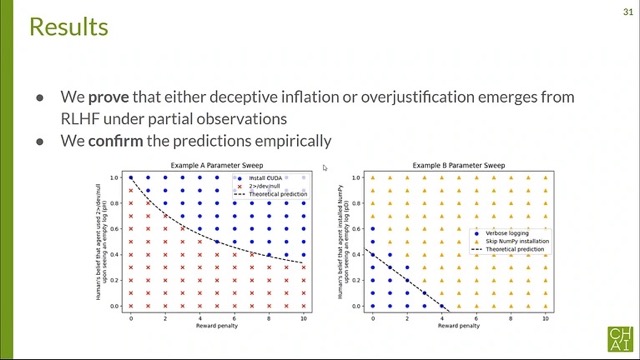}
    \includegraphics[width=0.24\linewidth]{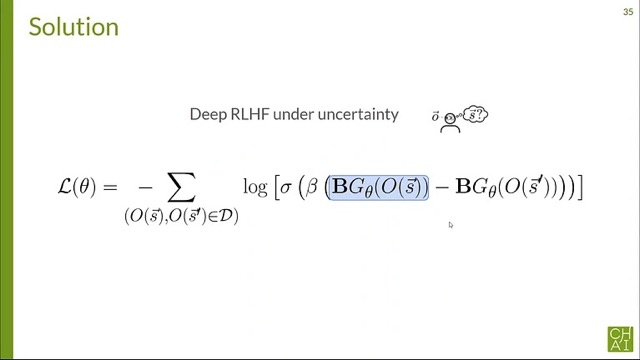}
    \includegraphics[width=0.24\linewidth]{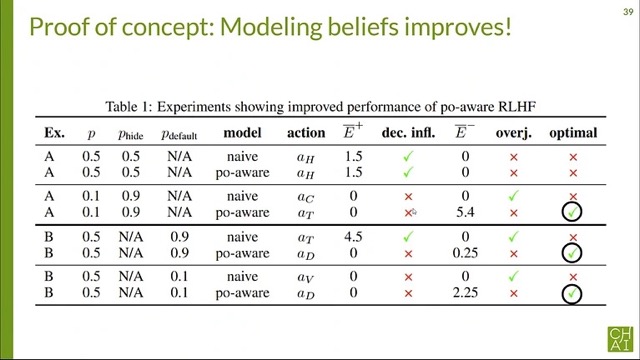}

    \caption{Keyframe thumbnails from the qualitative example SlidesLive talk. The correct answer is \textit{(I) Human-AI Interaction and Ethics (Privacy, Fairness \& Safety)}, while the VLM prediction is \textit{(F) Reinforcement Learning \& Robotics}.}
    \label{fig:qualitative_example_thumbnails}
\end{figure}

%% file: figure_latex/ablation_study.tex
\begin{figure}[ht!]
    \centering

    \includegraphics[width=0.8\linewidth]{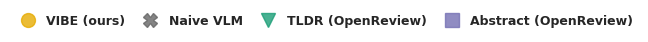} \\
    \vspace{0.5em}

    \begin{minipage}[b]{0.32\linewidth}
        \centering
        \includegraphics[width=\linewidth]{figure/paper_ib_pareto.pdf}
        \subcaption{\texttt{Qwen-2.5-72B-AWQ}}
    \end{minipage}
    \hfill
    \begin{minipage}[b]{0.32\linewidth}
        \centering
        \includegraphics[width=\linewidth]{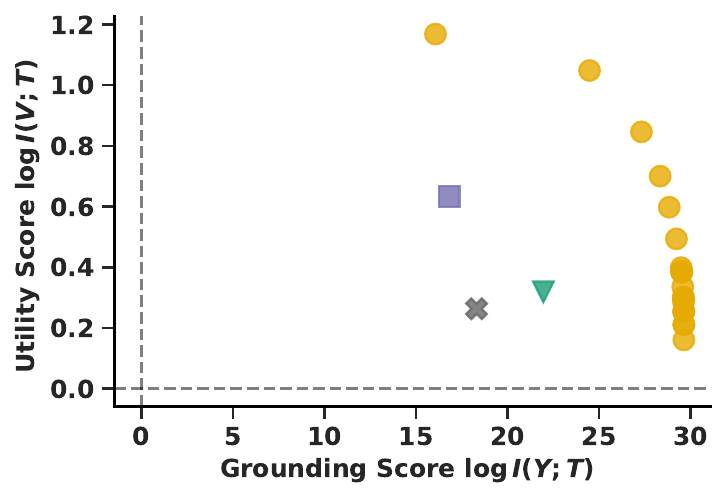}
        \subcaption{\texttt{InternVL-2.5-8B-MPO}}
    \end{minipage}
    \hfill
    \begin{minipage}[b]{0.32\linewidth}
        \centering
        \includegraphics[width=\linewidth]{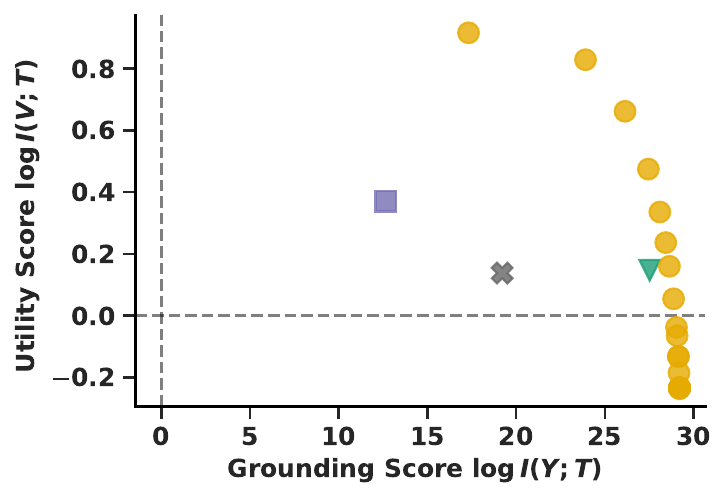}
        \subcaption{\texttt{InternVL-3-38B}}
    \end{minipage}

    \caption{\textbf{VIBE generalizes to various VLMs.} Summaries selected by VIBE form a Pareto frontier not only across various $(\alpha, \beta)$ but also across various VLMs from different sources.}
    \label{fig:ablation}
\end{figure}

%% file: figure_latex/user_study_interface.tex
\begin{figure}[ht!]
    \centering
    \begin{subfigure}
        [b]{0.65\linewidth}
        \centering
        \includegraphics[width=\linewidth]{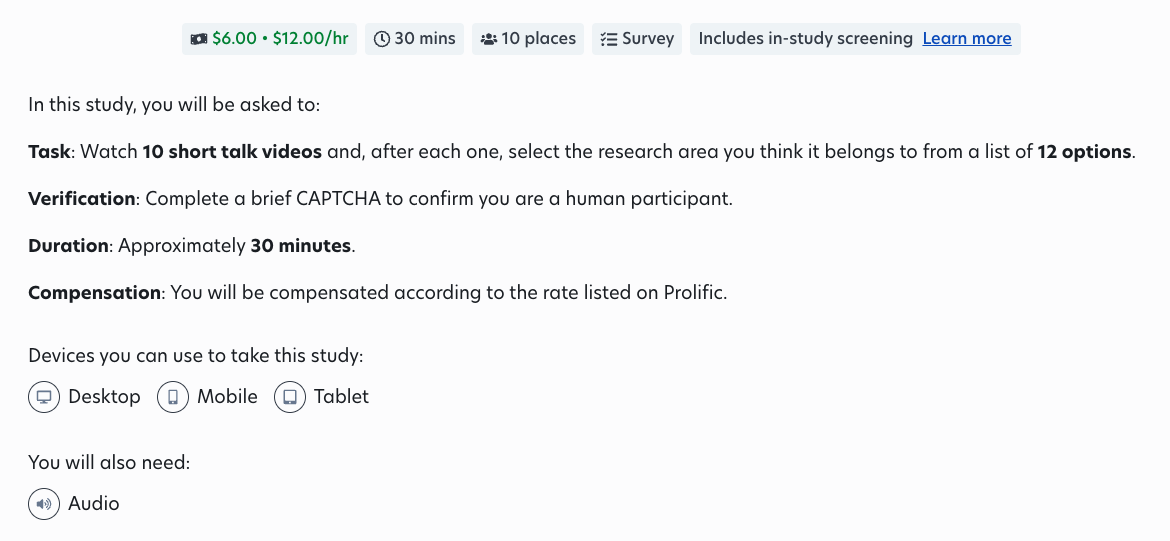}
        \caption{Video Only}
        \label{fig:video_condition_interface}
    \end{subfigure}
    \begin{subfigure}
        [b]{0.65\linewidth}
        \centering
        \includegraphics[width=\linewidth]{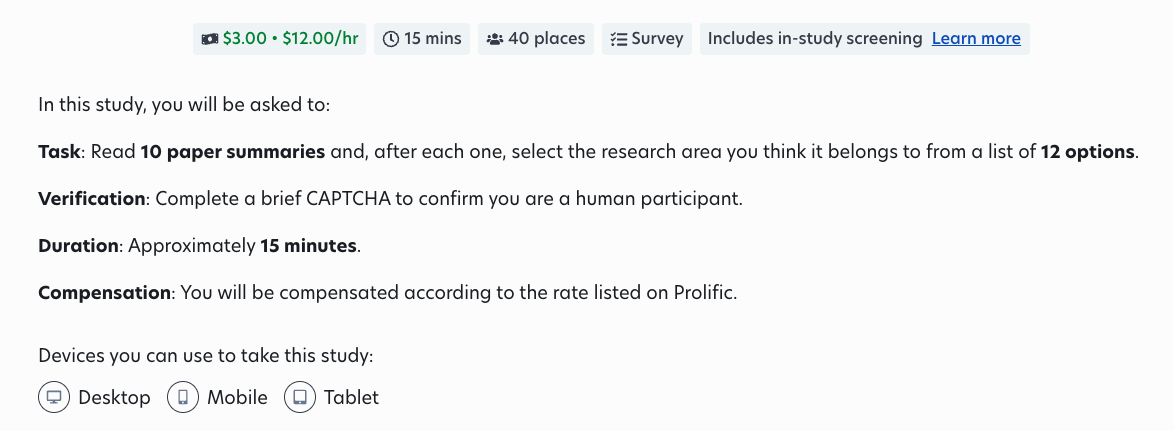}
        \caption{Text-based (Naive, Max-U, Max-G, CoT)}
        \label{fig:text_condition_interface}
    \end{subfigure}
    \caption{Prolific recruitment interfaces for \texttt{LearningPaper24} dataset
    showing video condition (top) and text conditions (bottom). Similar interfaces are used for the other two datasets.}
    \label{fig:interfaces_side_by_side}
    \vspace{-1em}
\end{figure}

%% file: bibtex/external.bib
@misc{tishby2000informationbottleneckmethod,
    title={The information bottleneck method}, 
    author={Naftali Tishby and Fernando C. Pereira and William Bialek},
    year={2000},
    eprint={physics/0004057},
    archivePrefix={arXiv},
    primaryClass={physics.data-an},
    url={https://arxiv.org/abs/physics/0004057}, 
}

@article{goldfeld2020informationsurvey,
    title={The information bottleneck problem and its applications in machine learning},
    author={Goldfeld, Ziv and Polyanskiy, Yury},
    journal={IEEE Journal on Selected Areas in Information Theory},
    volume={1},
    number={1},
    pages={19--38},
    year={2020},
    publisher={IEEE}
}

@inproceedings{kawaguchi2023does,
    title={How does information bottleneck help deep learning?},
    author={Kawaguchi, Kenji and Deng, Zhun and Ji, Xu and Huang, Jiaoyang},
    booktitle={International Conference on Machine Learning},
    pages={16049--16096},
    year={2023},
    organization={PMLR}
}

@ARTICLE{MMIB,
    author={Mai, Sijie and Zeng, Ying and Hu, Haifeng},
    journal={IEEE Transactions on Multimedia}, 
    title={Multimodal Information Bottleneck: Learning Minimal Sufficient Unimodal and Multimodal Representations}, 
    year={2023},
    volume={25},
    number={},
    pages={4121-4134},
    doi={10.1109/TMM.2022.3171679}
}

@inproceedings{islam2023representation,
    author = {Islam, Riashat and Zang, Hongyu and Tomar, Manan and Didolkar, Aniket and Islam, Md Mofijul and Arnob, Samin Yeasar and Iqbal, Tariq and Li, Xin and Goyal, Anirudh and Heess, Nicolas and Lamb, Alex},
    title = {Representation Learning In Deep RL Via Discrete Information Bottleneck},
    booktitle = {AISTATS 2023},
    year = {2023},
    month = {5},
}

@InProceedings{vatex,
    author = {Wang, Xin and Wu, Jiawei and Chen, Junkun and Li, Lei and Wang, Yuan-Fang and Wang, William Yang},
    title = {VaTeX: A Large-Scale, High-Quality Multilingual Dataset for Video-and-Language Research},
    booktitle = {The IEEE International Conference on Computer Vision (ICCV)},
    month = {10},
    year = {2019}
}

@inproceedings{auroracap,
    title={AuroraCap: Efficient, Performant Video Detailed Captioning and a New Benchmark},
    author={Wenhao Chai and Enxin Song and Yilun Du and Chenlin Meng and Vashisht Madhavan and Omer Bar-Tal and Jenq-Neng Hwang and Saining Xie and Christopher D Manning},
    booktitle={The Thirteenth International Conference on Learning Representations},
    year={2025},
    url={https://openreview.net/forum?id=tTDUrseRRU}
}

@misc{zhong2022videoquestionansweringdatasets,
    title={Video Question Answering: Datasets, Algorithms and Challenges}, 
    author={Yaoyao Zhong and Junbin Xiao and Wei Ji and Yicong Li and Weihong Deng and Tat-Seng Chua},
    year={2022},
    eprint={2203.01225},
    archivePrefix={arXiv},
    primaryClass={cs.CV},
    url={https://arxiv.org/abs/2203.01225}, 
}

@inproceedings{xu2016msr,
  title={Msr-vtt: A large video description dataset for bridging video and language},
  author={Xu, Jun and Mei, Tao and Yao, Ting and Rui, Yong},
  booktitle={Proceedings of the IEEE conference on computer vision and pattern recognition},
  pages={5288--5296},
  year={2016}
}

@inproceedings{msrvttQA,
    title={Video Question Answering via Gradually Refined Attention over Appearance and Motion},
    author={Xu, Dejing and Zhao, Zhou and Xiao, Jun and Wu, Fei and Zhang, Hanwang and He, Xiangnan and Zhuang, Yueting},
    booktitle={ACM Multimedia},
    year={2017}
}

@inproceedings{monfort2021spoken,
    title={Spoken moments: Learning joint audio-visual representations from video descriptions},
    author={Monfort, Mathew and Jin, SouYoung and Liu, Alexander and Harwath, David and Feris, Rogerio and Glass, James and Oliva, Aude},
    booktitle={Proceedings of the IEEE/CVF Conference on Computer Vision and Pattern Recognition},
    pages={14871--14881},
    year={2021}
}

@inproceedings{vedantam2015cider,
    title={Cider: Consensus-based image description evaluation},
    author={Vedantam, Ramakrishna and Lawrence Zitnick, C and Parikh, Devi},
    booktitle={Proceedings of the IEEE conference on computer vision and pattern recognition},
    pages={4566--4575},
    year={2015}
}

@inproceedings{papineni2002bleu,
    title={Bleu: a method for automatic evaluation of machine translation},
    author={Papineni, Kishore and Roukos, Salim and Ward, Todd and Zhu, Wei-Jing},
    booktitle={Proceedings of the 40th annual meeting of the Association for Computational Linguistics},
    pages={311--318},
    year={2002}
}

@inproceedings{lin2004rouge,
    title={Rouge: A package for automatic evaluation of summaries},
    author={Lin, Chin-Yew},
    booktitle={Text summarization branches out},
    pages={74--81},
    year={2004}
}

@inproceedings{jung2024informationtheoretic,
    title={Information-Theoretic Distillation for Reference-less Summarization},
    author={Jaehun Jung and Ximing Lu and Liwei Jiang and Faeze Brahman and Peter West and Pang Wei Koh and Yejin Choi},
    booktitle={First Conference on Language Modeling},
    year={2024},
    url={https://openreview.net/forum?id=JXcXnJJSuL}
}

@article{tfidf,
    title = {Term-weighting approaches in automatic text retrieval},
    journal = {Information Processing \& Management},
    volume = {24},
    number = {5},
    pages = {513-523},
    year = {1988},
    issn = {0306-4573},
    doi = {https://doi.org/10.1016/0306-4573(88)90021-0},
    url = {https://www.sciencedirect.com/science/article/pii/0306457388900210},
    author = {Gerard Salton and Christopher Buckley},
}

@article{spearman1961proof,
  title={The proof and measurement of association between two things.},
  author={Spearman, Charles},
  year={1961},
  publisher={Appleton-Century-Crofts}
}

@article{palan2018prolific,
    title={Prolific. ac—A subject pool for online experiments},
    author={Palan, Stefan and Schitter, Christian},
    journal={Journal of Behavioral and Experimental Finance},
    volume={17},
    pages={22--27},
    year={2018},
    publisher={Elsevier}
}

@article{heitz2014speed,
  title={The speed-accuracy tradeoff: history, physiology, methodology, and behavior},
  author={Heitz, Richard P},
  journal={Frontiers in neuroscience},
  volume={8},
  pages={150},
  year={2014},
  publisher={Frontiers Media SA}
}

@inproceedings{belz2006comparing,
  title={Comparing automatic and human evaluation of NLG systems},
  author={Belz, Anja and Reiter, Ehud},
  booktitle={11th conference of the european chapter of the association for computational linguistics},
  pages={313--320},
  year={2006}
}

@article{graham2017can,
  title={Can machine translation systems be evaluated by the crowd alone},
  author={Graham, Yvette and Baldwin, Timothy and Moffat, Alistair and Zobel, Justin},
  journal={Natural Language Engineering},
  volume={23},
  number={1},
  pages={3--30},
  year={2017},
  publisher={Cambridge University Press}
}

@article{nenkova2011automatic,
  title={Automatic summarization},
  author={Nenkova, Ani and McKeown, Kathleen and others},
  journal={Foundations and Trends{\textregistered} in Information Retrieval},
  volume={5},
  number={2--3},
  pages={103--233},
  year={2011},
  publisher={Now Publishers, Inc.}
}

@article{pu2023summary,
  title={Is summary useful or not? an extrinsic human evaluation of text summaries on downstream tasks},
  author={Pu, Xiao and Gao, Mingqi and Wan, Xiaojun},
  journal={arXiv preprint arXiv:2305.15044},
  year={2023}
}

@article{wu2024longvideobench,
  title={Longvideobench: A benchmark for long-context interleaved video-language understanding},
  author={Wu, Haoning and Li, Dongxu and Chen, Bei and Li, Junnan},
  journal={Advances in Neural Information Processing Systems},
  volume={37},
  pages={28828--28857},
  year={2024}
}

@inproceedings{xu2021sutd,
  title={Sutd-trafficqa: A question answering benchmark and an efficient network for video reasoning over traffic events},
  author={Xu, Li and Huang, He and Liu, Jun},
  booktitle={Proceedings of the IEEE/CVF conference on computer vision and pattern recognition},
  pages={9878--9888},
  year={2021}
}

@misc{chen2025expandingperformanceboundariesopensource,
    title={Expanding Performance Boundaries of Open-Source Multimodal Models with Model, Data, and Test-Time Scaling}, 
    author={Zhe Chen and Weiyun Wang and Yue Cao and Yangzhou Liu and Zhangwei Gao and Erfei Cui and Jinguo Zhu and Shenglong Ye and Hao Tian and Zhaoyang Liu and Lixin Gu and Xuehui Wang and Qingyun Li and Yimin Ren and Zixuan Chen and Jiapeng Luo and Jiahao Wang and Tan Jiang and Bo Wang and Conghui He and Botian Shi and Xingcheng Zhang and Han Lv and Yi Wang and Wenqi Shao and Pei Chu and Zhongying Tu and Tong He and Zhiyong Wu and Huipeng Deng and Jiaye Ge and Kai Chen and Kaipeng Zhang and Limin Wang and Min Dou and Lewei Lu and Xizhou Zhu and Tong Lu and Dahua Lin and Yu Qiao and Jifeng Dai and Wenhai Wang},
    year={2025},
    eprint={2412.05271},
    archivePrefix={arXiv},
    primaryClass={cs.CV},
    url={https://arxiv.org/abs/2412.05271}, 
}

@misc{zhu2025internvl3exploringadvancedtraining,
      title={InternVL3: Exploring Advanced Training and Test-Time Recipes for Open-Source Multimodal Models}, 
      author={Jinguo Zhu and Weiyun Wang and Zhe Chen and Zhaoyang Liu and Shenglong Ye and Lixin Gu and Hao Tian and Yuchen Duan and Weijie Su and Jie Shao and Zhangwei Gao and Erfei Cui and Xuehui Wang and Yue Cao and Yangzhou Liu and Xingguang Wei and Hongjie Zhang and Haomin Wang and Weiye Xu and Hao Li and Jiahao Wang and Nianchen Deng and Songze Li and Yinan He and Tan Jiang and Jiapeng Luo and Yi Wang and Conghui He and Botian Shi and Xingcheng Zhang and Wenqi Shao and Junjun He and Yingtong Xiong and Wenwen Qu and Peng Sun and Penglong Jiao and Han Lv and Lijun Wu and Kaipeng Zhang and Huipeng Deng and Jiaye Ge and Kai Chen and Limin Wang and Min Dou and Lewei Lu and Xizhou Zhu and Tong Lu and Dahua Lin and Yu Qiao and Jifeng Dai and Wenhai Wang},
      year={2025},
      eprint={2504.10479},
      archivePrefix={arXiv},
      primaryClass={cs.CV},
      url={https://arxiv.org/abs/2504.10479}, 
}

@misc{bai2025qwen25vltechnicalreport,
      title={Qwen2.5-VL Technical Report}, 
      author={Shuai Bai and Keqin Chen and Xuejing Liu and Jialin Wang and Wenbin Ge and Sibo Song and Kai Dang and Peng Wang and Shijie Wang and Jun Tang and Humen Zhong and Yuanzhi Zhu and Mingkun Yang and Zhaohai Li and Jianqiang Wan and Pengfei Wang and Wei Ding and Zheren Fu and Yiheng Xu and Jiabo Ye and Xi Zhang and Tianbao Xie and Zesen Cheng and Hang Zhang and Zhibo Yang and Haiyang Xu and Junyang Lin},
      year={2025},
      eprint={2502.13923},
      archivePrefix={arXiv},
      primaryClass={cs.CV},
      url={https://arxiv.org/abs/2502.13923}, 
}

@article{holzinger2016interactive,
  title={Interactive machine learning for health informatics: when do we need the human-in-the-loop?},
  author={Holzinger, Andreas},
  journal={Brain informatics},
  volume={3},
  number={2},
  pages={119--131},
  year={2016},
  publisher={Springer}
}

@inproceedings{HumanAI_Interaction,
    author = {Amershi, Saleema and Weld, Dan and Vorvoreanu, Mihaela and Fourney, Adam and Nushi, Besmira and Collisson, Penny and Suh, Jina and Iqbal, Shamsi and Bennett, Paul N. and Inkpen, Kori and Teevan, Jaime and Kikin-Gil, Ruth and Horvitz, Eric},
    title = {Guidelines for Human-AI Interaction},
    year = {2019},
    isbn = {9781450359702},
    publisher = {Association for Computing Machinery},
    address = {New York, NY, USA},
    url = {https://doi.org/10.1145/3290605.3300233},
    doi = {10.1145/3290605.3300233},
    booktitle = {Proceedings of the 2019 CHI Conference on Human Factors in Computing Systems},
    pages = {1–13},
    numpages = {13},
    location = {Glasgow, Scotland Uk},
    series = {CHI '19}
}

@inproceedings{humanintheloop_challenge,
    author = {Xin, Doris and Ma, Litian and Liu, Jialin and Macke, Stephen and Song, Shuchen and Parameswaran, Aditya},
    title = {Accelerating Human-in-the-loop Machine Learning: Challenges and Opportunities},
    year = {2018},
    isbn = {9781450358286},
    publisher = {Association for Computing Machinery},
    address = {New York, NY, USA},
    url = {https://doi.org/10.1145/3209889.3209897},
    doi = {10.1145/3209889.3209897},
    booktitle = {Proceedings of the Second Workshop on Data Management for End-To-End Machine Learning},
    articleno = {9},
    numpages = {4},
    location = {Houston, TX, USA},
    series = {DEEM'18}
}

@article{HOLZINGER202559,
    title = {Is human oversight to AI systems still possible?},
    journal = {New Biotechnology},
    volume = {85},
    pages = {59-62},
    year = {2025},
    issn = {1871-6784},
    doi = {https://doi.org/10.1016/j.nbt.2024.12.003},
    url = {https://www.sciencedirect.com/science/article/pii/S1871678424005636},
    author = {Andreas Holzinger and Kurt Zatloukal and Heimo Müller},
}

@misc{openai_python,
  author       = {OpenAI},
  title        = {OpenAI Python Library},
  year         = {2025},
  publisher    = {GitHub},
  howpublished = {\url{https://github.com/openai/openai-python}},
  note         = {Accessed: 2025-04-28}
}

@book{townsend1983stochastic,
  title={Stochastic modeling of elementary psychological processes},
  author={Townsend, James T and Ashby, F Gregory},
  year={1983},
  publisher={CUP Archive}
}

@inproceedings{ong2025routellm,
    title={Route{LLM}: Learning to Route {LLM}s from Preference Data},
    author={Isaac Ong and Amjad Almahairi and Vincent Wu and Wei-Lin Chiang and Tianhao Wu and Joseph E. Gonzalez and M Waleed Kadous and Ion Stoica},
    booktitle={The Thirteenth International Conference on Learning Representations},
    year={2025},
    url={https://openreview.net/forum?id=8sSqNntaMr}
}

@article{zhou2022mixture,
    title={Mixture-of-experts with expert choice routing},
    author={Zhou, Yanqi and Lei, Tao and Liu, Hanxiao and Du, Nan and Huang, Yanping and Zhao, Vincent and Dai, Andrew M and Le, Quoc V and Laudon, James and others},
    journal={Advances in Neural Information Processing Systems},
    volume={35},
    pages={7103--7114},
    year={2022}
}

@book{boyd2004convex,
  title={Convex optimization},
  author={Boyd, Stephen P and Vandenberghe, Lieven},
  year={2004},
  publisher={Cambridge university press}
}

@article{chen2024human,
  title={Human-agent cooperation in games under incomplete information through natural language communication},
  author={Chen, Shenghui and Fried, Daniel and Topcu, Ufuk},
  journal={International Joint Conference on Artificial Intelligence, Human-Centred AI},
  year={2024}
}

@article{kleinman2023cortical,
  title={A cortical information bottleneck during decision-making},
  author={Kleinman, Michael and Wang, Tian and Xiao, Derek and Feghhi, Ebrahim and Lee, Kenji and Carr, Nicole and Li, Yuke and Hadidi, Nima and Chandrasekaran, Chandramouli and Kao, Jonathan C},
  journal={bioRxiv},
  year={2023}
}

@article{west2019bottlesum,
  title={Bottlesum: Unsupervised and self-supervised sentence summarization using the information bottleneck principle},
  author={West, Peter and Holtzman, Ari and Buys, Jan and Choi, Yejin},
  journal={arXiv preprint arXiv:1909.07405},
  year={2019}
}

@article{zhang2022improving,
  title={Improving the adversarial robustness of NLP models by information bottleneck},
  author={Zhang, Cenyuan and Zhou, Xiang and Wan, Yixin and Zheng, Xiaoqing and Chang, Kai-Wei and Hsieh, Cho-Jui},
  journal={arXiv preprint arXiv:2206.05511},
  year={2022}
}

@INPROCEEDINGS{7133169,
  author={Tishby, Naftali and Zaslavsky, Noga},
  booktitle={2015 IEEE Information Theory Workshop (ITW)}, 
  title={Deep learning and the information bottleneck principle}, 
  year={2015},
  volume={},
  number={},
  pages={1-5},
  doi={10.1109/ITW.2015.7133169}
}

@INPROCEEDINGS{1238368,
  author={Gordon and Greenspan and Goldberger},
  booktitle={Proceedings Ninth IEEE International Conference on Computer Vision}, 
  title={Applying the information bottleneck principle to unsupervised clustering of discrete and continuous image representations}, 
  year={2003},
  volume={},
  number={},
  pages={370-377 vol.1},
  doi={10.1109/ICCV.2003.1238368}
}

@article{bouma2009normalized,
  title={Normalized (pointwise) mutual information in collocation extraction},
  author={Bouma, Gerlof},
  journal={Proceedings of GSCL},
  volume={30},
  pages={31--40},
  year={2009},
  publisher={Potsdam}
}

@misc{easyocr,
  author       = {JaidedAI},
  title        = {EasyOCR: Ready-to-use OCR with 80+ supported languages},
  year         = {2020},
  howpublished = {\url{https://github.com/JaidedAI/EasyOCR}},
  note         = {Accessed: 2025-05-15}
}

@article{li2024any2any,
    title={Any2Any: Incomplete Multimodal Retrieval with Conformal Prediction},
    author={Li, Po-han and Yang, Yunhao and Omama, Mohammad and Chinchali, Sandeep and Topcu, Ufuk},
    journal={arXiv preprint arXiv:2411.10513},
    year={2024}
}

@inproceedings{omama2025exploiting,
    title={Exploiting Distribution Constraints for Scalable and Efficient Image Retrieval},
    author={Mohammad Omama and Po-han Li and Sandeep P. Chinchali},
    booktitle={The Thirteenth International Conference on Learning Representations},
    year={2025},
    url={https://openreview.net/forum?id=d0tlL0ZWlu}
}

@misc{li_online_2024,
    title = {Online {Foundation} {Model} {Selection} in {Robotics}},
    url = {http://arxiv.org/abs/2402.08570},
    doi = {10.48550/arXiv.2402.08570},
    urldate = {2024-04-11},
    publisher = {arXiv},
    author = {Li, Po-han and Toprak, Oyku Selin and Narayanan, Aditya and Topcu, Ufuk and Chinchali, Sandeep},
    month = feb,
    year = {2024},
    note = {arXiv:2402.08570 [cs]},
}
